\documentclass[twocolumn,letterpaper]{IEEEAerospaceCLS}  

\usepackage[]{graphicx}    
\usepackage[colorlinks=false, urlcolor=blue, linkcolor=black, citecolor=black]{hyperref} 

\usepackage{multirow}  

\usepackage{cleveref}
\usepackage{xcolor} 
\usepackage{cite} 
\usepackage{algorithm} 
\usepackage[noend]{algpseudocode} 
\newcommand\highlight[1]{\textcolor{blue}{#1}} 
\newcommand\highlightgreen[1]{\textcolor{violet}{#1}} %

\normalsize 
\usepackage{booktabs} 

\usepackage{subcaption} %

\newcommand{\ignore}[1]{}  
\begin{document}
\title{Bringing Federated Learning to Space}

\author{Grace Ra Kim\\ 
Stanford University\\
496 Lomita Mall \\
Stanford, CA 94305 \\
gkim65@stanford.edu, 
\and 
Filip Svoboda, Nicholas D. Lane\\
University of Cambridge\\
15 JJ Thomson Avenue\\
Cambridge, CB3 0FD\\
fs437@cam.ac.uk,  
ndl32@cam.ac.uk
\thanks{\footnotesize 979-8-3315-7360-7/26/$\$31.00$ \copyright2026 IEEE}              
}

\maketitle

\thispagestyle{plain}
\pagestyle{plain}

\maketitle

\thispagestyle{plain}
\pagestyle{plain}

\begin{abstract}
\textbf{
As Low Earth Orbit (LEO) satellite constellations rapidly expand to hundreds and thousands of spacecraft, the need for distributed on-board machine learning becomes critical to address downlink bandwidth limitations. Federated learning (FL) offers a promising framework to conduct collaborative model training across satellite networks. Realizing its benefits in space naturally requires addressing space-specific constraints, from intermittent connectivity to dynamics imposed by orbital motion. This work presents the first systematic feasibility analysis of adapting off-the-shelf FL algorithms for satellite constellation deployment. We introduce a comprehensive “space-ification” framework that adapts terrestrial algorithms (FedAvg, FedProx, FedBuff) to operate under orbital constraints, producing an orbital-ready suite of FL algorithms. We then evaluate these space-ified methods through extensive parameter sweeps across 768 constellation configurations that vary cluster sizes (1–10), satellites per cluster (1–10), and ground station networks (1–13). Our analysis demonstrates that space-adapted FL algorithms efficiently scale to constellations of up to 100 satellites, achieving performance close to the centralized ideal. Multi-month training cycles can be reduced to days, corresponding to a 9X speedup through orbital scheduling and local coordination within satellite clusters. These results provide actionable insights for future mission designers, enabling distributed on-board learning for more autonomous, resilient, and data-driven satellite operations.}
\end{abstract} 
\vspace{-1em}

\tableofcontents
\section{Introduction}
Low Earth Orbit (LEO) satellite constellations are expanding rapidly, supporting applications in Earth observation (EO), telecommunications, and navigation. Large-scale constellations such as Planet Labs’ Dove fleet, SpaceX’s Starlink, and Amazon’s Project Kuiper already consist of hundreds to thousands of spacecraft, representing some of the largest distributed systems ever deployed. This unprecedented scale is driving a dramatic increase in the volume and diversity of space-based data.

Earth observation missions in particular bear the brunt of this data challenge. High-resolution missions such as Landsat-8 produce $1.8$ GB per scene and more than $400$ TB annually~\cite{wulder_global_2016}. At constellation scale, Planet Labs’ fleet of over $200$ satellites generates terabytes of imagery each day \cite{PlanetLabs_PSSceneImagerySpec_2023}. These massive datasets enable applications ranging from agriculture monitoring~\cite{nguyen_monitoring_2020} and disease prediction~\cite{ford_using_2009} to disaster response~\cite{visser_real-time_2004}. However, their value depends on timely downlink and processing, especially in time-sensitive contexts such as emergency management and extreme weather forecasting~\cite{giggenbach_optical_2009}.

Despite advances in space-based sensing, opportunities to transmit data to the ground remain limited. Depending on orbital parameters and ground coverage, contact opportunities can range from $30$ minutes to $9$ hours~\cite{larson_space_1999}, and current networks may downlink only a fraction of collected imagery. For example, current ground station networks are only able to receive $\sim2\%$ of Landsat-8's hyperspectral image frames during a single orbit~\cite{denby_kodan_2023}. This mismatch between data generation and downlink capacity creates a fundamental bottleneck: satellites must either store large volumes of data onboard, constrained by limited memory resources, or discard potentially valuable information.

Several approaches have been proposed to mitigate this bottleneck. One approach suggests relay architectures, such as using Starlink satellites for data forwarding, to accelerate data delivery to the ground~\cite{wang_enhancing_2022}. Others have pushed for onboard processing methods, including machine learning (ML) for image pre-selection, to reduce downlink demand by transmitting only high-value observations. For instance, the RaVÆN model was executed directly on-board the ION SCV004 spacecraft to detect extreme natural events for prioritized downlink~\cite{ruuvzivcka2023fast}. While promising, these methods address only part of the challenge. The increasing availability of edge computing capabilities in modern satellites enables sophisticated in-orbit data processing, opening the possibility for collaborative training across distributed spacecraft. By training machine learning (ML) models directly onboard, satellites can process observations immediately after collection, reducing the need to downlink raw data and alleviating traditional transmission bottlenecks.

Federated learning (FL), developed for distributed edge devices, offers a compelling framework for this challenge~\cite{matthiesen2023federated}. In FL, each client trains locally on its own data and shares only model updates, rather than raw observations. Applied to satellite constellations, FL would allow spacecraft to immediately process and learn from their observations while exchanging minimal information, reducing reliance on sparse ground contacts. However, deploying FL in space introduces unique challenges: intermittent connectivity, variable communication delays, and limited onboard computational resources, that require adapting terrestrial FL methods for effective use in orbit.

This work introduces methods for adapting terrestrial FL algorithms to satellite constellations, addressing the core challenge of distributed learning under space-specific constraints. Our contributions are threefold. We provide a (1) \textbf{space-ification framework}: A modular process for adapting any FL algorithm to orbital conditions, including client selection, round completion, and model evaluation. We demonstrate this with \textit{FedAvg}, \textit{FedProx}, and \textit{FedBuff}. (2) \textbf{Performance augmentations}: Two constellation-specific enhancements, scheduling optimizations and intra-satellite communications, that can be introduced into any space-ified FL algorithm to reduce total round durations, idle times, and improve resource utilization. (3) \textbf{Extensive experimental testing}: Simulations across 768 configurations of ground station networks, cluster counts, and satellite densities, offering actionable insights for collaborative on-orbit learning. Using these methods, we show that existing terrestrial FL algorithms, once adapted through our space-ification framework, can be effectively deployed across satellite constellations. All adapted algorithms achieve $>80\%$ accuracy on standard benchmarks, and with our performance augmentations, training times can be reduced by up to $9$X (from three months to $\sim10$ days) for constellations of 100 satellites. These results demonstrate that conventional FL techniques, with appropriate modifications, can support scalable, collaborative on-orbit learning in real-world space environments.

\section{Related Work}
Training machine learning models directly on-board satellites has become a critical area of research for enabling autonomous and collaborative space operations. Prior work in this domain can be broadly categorized into two directions: the development of ML models suitable for on-board training and inference, and the adaptation of federated learning protocols to the unique constraints of satellite constellations.

\subsection{On-Board ML Training for Satellites}

On-board machine learning (ML) training has emerged as a promising approach for enabling autonomous satellite operations. Early efforts include RaVÆN, an unsupervised change detection model that enabled local event identification on satellites \cite{ruuvzivcka2023fast}, and $\Phi$-Sat-1, which demonstrated cloud filtering via deep neural networks, reducing unnecessary downlink \cite{ghasemi2025onboard}. The WorldFloods payload further established the feasibility of in-orbit model retraining and agile adaptation to sensor drift, providing evidence for robust satellite learning capabilities \cite{mateo2023orbit}. OPS-SAT extended these advances by supporting competitive, collaborative onboard ML workflows for image classification and anomaly detection, confirming that a range of models can operate autonomously and efficiently under real space constraints \cite{meoni2024ops}. Together, these missions represent some of the first demonstrations of training ML models directly onboard spacecraft, thereby establishing flight heritage and validating the feasibility of on-orbit learning. Since satellites often operate for extended periods without ground contact, onboard learning is particularly valuable for enabling collaborative model training. However, these efforts remain limited to single-satellite operation and do not address scaling to distributed constellations, leaving open questions about multi-satellite collaboration and coordinated learning across spaceborne clients.

\subsection{Federated Learning}

While on-board ML enables individual satellites to process observations and make autonomous decisions, many applications such as disaster monitoring across wide regions would benefit from collaboration across multiple satellites. Federated learning provides a natural framework for this type of distributed, privacy-preserving model training, allowing satellites to collectively improve a global model without sharing raw data~\cite{mcmahan_communication-efficient_2016}. In FL, each client (ranging from mobile devices to data centers or satellites) trains a model on its local dataset and periodically shares only the learned updates with a central server, which aggregates them to update the global model. By exchanging model parameters rather than raw data, FL preserves privacy and supports training across networks with limited or intermittent connectivity~\cite{mothukuri_survey_2021, matthiesen2023federated, fan_fate-llm_2023}. Foundational algorithms such as \textit{FedAvg}~\cite{mcmahan_communication-efficient_2016} have been extended with methods like \textit{FedProx} and \textit{FedBuff} to address challenges including uneven data distributions across clients, communication overhead, and heterogeneous computational capabilities~\cite{li_federated_2020, nguyen2022federated}. While terrestrial FL provides valuable design principles, many of its underlying assumptions such as on-demand client-server communication, random client sampling, and low-latency, and reliable networks do not hold in space. Adapting FL for satellite constellations therefore requires careful consideration of orbital dynamics, limited connectivity, and constrained computational resources.

\subsection{Federated Learning in Satellite Constellations}
Redesigning traditional federated learning algorithms to satellite constellations introduces unique challenges that do not arise in terrestrial deployments. Unlike terrestrial networks, satellite constellations operate in environments where contact times with ground stations are brief, satellite trajectories are highly deterministic, and the number of participating satellites is relatively small, increasing the significance of each agent in the learning process. These characteristics create both opportunities and constraints: predictable orbits allow for optimized scheduling of communication, while limited connectivity and constrained computational resources restrict the direct adoption of terrestrial FL protocols.

Existing research has begun to address these challenges, but often in a fragmented manner. Prior works have exploited deterministic orbits for optimized communication scheduling~\cite{razmi_ground-assisted_2022, razmi_2022_scheduling}, analyzed the potential of inter-satellite links to support distributed learning~\cite{zhai_fedleo_2024}, and explored the use of High-Altitude Platforms (HAPs) to enhance connectivity~\cite{elmahallawy_fedhap_2022, elmahallawy_communication-efficient_2024}. While these contributions mark important first steps toward adapting FL to the space environment, they typically examine narrow scenarios such as small-scale constellations, limited benchmark problems, or simplified communication architectures. There is no unifying framework that systematically integrates the full range of satellite operational constraints into the FL algorithm design and deployment.

To date, no study has provided both a broad analysis of how terrestrial FL assumptions break down in the context of orbital systems and a comprehensive methodology for adapting any FL algorithm to satellite networks. Critical aspects, including the implications of deterministic motion, intermittent connectivity, scalability to large constellations, and operational orbital constraints, remain underexplored. This work addresses these gaps by presenting a unified framework for space-adapting terrestrial FL algorithms, alongside a detailed feasibility study demonstrating their practical and scalable deployment across large satellite constellations.
\section{Space-ification of Federated Learning Algorithms}

To apply FL in space, algorithms must address a distinct set of hardware and communication constraints that differ from traditional FL on edge device deployment. By accounting for the operational constraints inherent to the orbital environment, any FL algorithm can undergo a ``space-ification'' process for deployment in space. In this section, we introduce our modular and algorithm-agnostic space-ification framework, designed to standardize the adaptation of terrestrial FL algorithms under realistic orbital conditions. While we illustrate this process concretely by rebuilding \textit{FedAvg}, the framework is broadly applicable and extensible to other FL algorithms. 
Similar procedures are applied to \textit{FedProx} and \textit{FedBuff}, algorithms particularly suited to the communication constraints of satellite constellations. Together, these space-ified aggregation strategies constitute the first standardized suite of terrestrial FL algorithms for orbital deployment.

\subsection{Mapping Federated Learning Concepts to Space}
Federated learning is a distributed machine learning paradigm where multiple clients collaboratively train a shared global model without directly exchanging raw data. While originally developed for terrestrial edge devices, its core concepts—such as client-server interactions, iterative training rounds, and model aggregation—can be adapted for satellite constellations by carefully incorporating the unique characteristics of the orbital environment.

Consider a constellation of Earth observation satellites, where each satellite leverages its onboard data to train a local deep neural network. In the FL framework, the constellation forms a distributed network of \textit{clients} $K$, indexed by $k$, each performing local training on their private datasets. Satellites serve as the clients within this protocol. The \textit{server}, which aggregates client updates into a global model, corresponds to the ground station network $G$, indexed by $g$, responsible for periodically receiving local model parameters and disseminating updated global parameters. Global coordination is maintained exclusively through this server, and direct communication among satellites (clients) is not required \cite{mcmahan_communication-efficient_2016}.

The training process proceeds in \textit{rounds}, indexed in timesteps $t$. In each round, the server transmits the current global model $w_t$ to the participating satellites. Each satellite then performs local stochastic gradient descent, via the function \textbf{ClientUpdate}($k,w$) on its onboard dataset $P_k$, producing its updated local model parameters $w_{t+1}^k$. These parameters are transmitted back to the ground station, which aggregates them according to a chosen strategy. For example in \textit{FedAvg}, which computes a weighted average of the local model updates based on dataset sizes, the global model is updated as
\begin{equation}
w_{t+1} \leftarrow \sum_{k \in S_t} \frac{n_k}{m_t} w^k_{t+1}, \quad m_t = \sum_{k\in S_t}n_k 
\end{equation}
where $S_t$ denotes the set of participating satellites in round $t$, $n_k$ is the dataset size of client $k$, and $m_t$ is the total data size across participating clients.

There are three key differences in applying FL in space compared to traditional terrestrial scenarios. First, communication windows between clients and the server are limited. Satellites have brief contact times with ground stations, typically 5-15 minutes every 90-180 minutes, eliminating the assumption of on-demand communication. Second, timing of contact windows for clients is deterministic but heterogeneous. While orbital mechanics are predictable, satellites experience different contact patterns, resulting in non-uniform client availability. Third, the number of clients is several orders of magnitude smaller than terrestrial distributed systems, resulting in proportionally larger contributions from each client in the model aggregation process.

Each of these constraints motivates modifications to standard FL protocols, which can incorporate further complexity for applying standardized distributed machine learning techniques on satellite constellations. However, the unique problem space, such as knowing each agent's trajectories and contact times, can be exploited to benefit the collaborative process of the group machine learning task. Together, these factors motivate the space-ification of FL algorithms, enabling effective operation under the deterministic and resource-constrained conditions of orbital environments. 

\begin{algorithm}[!ht]
\caption{\textit{\highlight{FedAvgSat}}. \textit{Base algorithm referenced from the original \textit{FedAvg} paper~\cite{mcmahan_communication-efficient_2016}, with modifications for satellite specific operations in \highlight{blue}.} \newline $K$ satellites indexed by $k$, $C$ max clients per round, $B$ local minibatch size, $E$ number of local epochs, $\mu$ learning rate, $P_k$ data onboard client $k$, and ground network $G$ indexed by $g$.}\label{alg:fedAvgSat}
\begin{algorithmic}
\algrenewcommand\algorithmicrequire{\textbf{Main server executes:}}
\algrenewcommand\algorithmicensure{\textbf{ClientUpdate($k$,$w$):}}
\Require 
    \State initialize $w_0$ (global model weights)
\For{each round $t = 0,1,2,...$} 
    \State{$c \leftarrow$ \highlight{min$(C, K)$}} 
    \State{$S_t \leftarrow $\highlight{(first $c$ idle clients that contact $G$)}}
    \State{\texttt{Send} $w_t$ to each $k$ in $S_t$}
    \For{each client $k \in S_t$ \textbf{in parallel}}
        \State{$w^k_t$ =  \textbf{ClientUpdate} $(k,w_t)$}
        \State{\highlight{\textit{Wait} for client $k$ to contact $G$ again after training}}
        \State{$w^k_{t+1} \leftarrow$ $w^k_t$}
    \EndFor
    \State{$m_t \leftarrow \Sigma_{k \in S_t} n_k$}
    \State{$w_{t+1} \leftarrow \Sigma_{k \in S_t} \frac{n_k}{m_t} w^k_{t+1}$}
\EndFor
\newline
\Ensure \Comment{\textit{Run on client/satellite k}}
\State {\highlight{\textit{Receive}  $w$ from ground station}}
\State{$\beta \leftarrow$ (Split $P_k$ into batches of size $B$)}
\For{ each local epoch $i$ from 1 to $E$}
\For{ batch $b \in \beta$ }
\State{$w \leftarrow w -\mu \nabla l(w;b)$}
\EndFor
\EndFor
\State{\highlight{Wait until reach nearest station in $G$, then return $w$}}
\end{algorithmic}
\end{algorithm}

\subsection{Space-ification Framework, Rebuilding \textit{FedAvg}} 
\label{sec:spacefiy}

To provide an accessible and familiar illustration of our framework, we detail the space-ification of the original FL algorithm \textit{FedAvg}. This serves as a baseline implementation and demonstrates the modular principles for adapting client selection, round completion, and evaluation stage protocols under satellite-specific constraints. Importantly, the framework is fully extensible: any terrestrial FL algorithm can be adapted using the same modular steps, allowing researchers to implement and test alternative algorithms within orbital environments. 

When performing the space-ification of an FL algorithm, three main considerations arise. (1) \textbf{Training Stage Client Selection:} For space-based settings, client selection cannot be randomized due to limited communication windows; every available contact must be taken advantage of. Consequently, clients are selected based on the first $C$ idle clients that make contact with a ground station. If the number $C$ is larger than the number of satellites on hand $K$, the minimum of $C$ and $K$ is taken. (2) \textbf{Round Completion:} Because of sparse communication opportunities, an FL round is considered complete only after every selected client in $S_t$ has returned its model parameters. If the server cannot prioritize clients with faster return times (an augmentation discussed in \Cref{sec:augmentations}), the algorithm must wait until all clients included in the round have submitted their updates before aggregating the global model. (3) \textbf{Evaluation Stage Client Selection:} The clients that initially participate in training may differ from those evaluating the aggregated model. Evaluation client selection follows the same protocol as training, which has a more pronounced effect in larger constellations where the total number of clients exceeds the predefined constant $C$. Applying the space-ification process to \textit{FedAvg} yields the algorithm \textit{FedAvgSat}, as illustrated in \Cref{alg:fedAvgSat}. 

\begin{algorithm}[!ht]
\caption{\textit{\highlight{FedProxSat}}. \textit{The base algorithm in black is referenced from the original \textit{FedProx} paper \cite{li_federated_2020}, with modifications for satellite specific operations in \highlight{blue}. \textit{FedProx} specific changes are in \highlightgreen{purple}.}\newline
$K$ satellites indexed by $k$, $C$ max clients per round, $B$ local minibatch size, $E$ number of local epochs, $\mu$ learning rate, $P_k$ data onboard client $k$, and ground network $G$ indexed by $g$.}\label{alg:fedProxSat}
\begin{algorithmic}
\algrenewcommand\algorithmicrequire{\textbf{Main Server executes:}}
\algrenewcommand\algorithmicensure{\textbf{ClientUpdate($k$,$w$):}}
\Require 
    \State initialize $w_0$ (global model weights)
\For{each round $t = 0,1,2,...$} 
    \State{$c \leftarrow$ \highlight{min$(C, K)$}} 
    \State{$S_t \leftarrow $\highlight{(first $c$ clients that contact $G$)}}
    \State{\textbf{Send} $w_t$ to each $k$ in $S_t$}
    \For{each client $k \in S_t$ \textbf{in parallel}}
        \State{$w^k_t$ =  \textbf{ClientUpdate} $(k,w_t)$}
        \State{\highlight{\textbf{Wait} for $k$ to contact $G$, client $k$ continues to train until contact is made}}
        \State{$w^k_{t+1} \leftarrow$ $w^k_t$}
    \EndFor
    \State{$m_t \leftarrow \Sigma_{k \in S_t} n_k$}
    \State{$w_{t+1} \leftarrow \Sigma_{k \in S_t} \frac{n_k}{m_t} w^k_{t+1}$}
\EndFor
\newline

\Ensure \Comment{\textit{Run on client/satellite k}}
\State {\highlight{\textbf{Receive}  $w$ from ground station}}
\State{$\beta \leftarrow$ (Split $P_k$ into batches of size $B$)}
\While{\highlightgreen{no access to ground station}}
\For{ batch $b \in \beta$ }
\State{$w_t = \mu \nabla l(w;b)$}
\State{$w \leftarrow w_t + \frac{\mu}{2} || w_t - w ||^2$} \Comment{\textit{\highlightgreen{Proximal Term, limits impact of local updates}}}
\EndFor
\EndWhile
\State{\highlight{return $w$ to nearest ground station}}
\end{algorithmic}
\end{algorithm}

\subsection{Space-ification of \textit{FedProx}: Allowing for Partial Updates}
Despite the cyclic nature of satellite orbits, ground station revisit times vary across satellites due to differences in assigned orbits and ground station locations. As a result, perfect synchronization for client parameter aggregation is not achievable. To address this, the \textit{FedProx} algorithm was selected to include as another terrestrial FL algorithm to undergo space-ification, as it allows incorporation of partial client updates. By leveraging this partial update scheme, client idle time is reduced, enabling more flexible participation in each FL round. Although FedProx is a synchronous FL method, it accommodates clients with less optimal orbital schedules by allowing them to perform variable amounts of local training. Compared to standard FL algorithms such as \textit{FedAvg}, clients are no longer required to complete a fixed number of training epochs $E$ in each round.

The space-ification modifications applied to \textit{FedAvg} are similarly applied to \textit{FedProx}. Client selection follows the same protocol as in FedAvg, without additional scheduling considerations. The full protocol is outlined in \Cref{alg:fedProxSat}. The primary difference lies in the client update function: each client trains for as many local epochs as possible rather than a server-defined fixed number. Local model updates are regularized via a proximal term that penalizes deviation from the global parameters provided at the start of the round.

\begin{algorithm} [!ht]
\caption{\textit{\highlight{FedBuffSat}}. \textit{The base algorithm in black is referenced from the original \textit{FedBuff} paper \cite{nguyen2022federated}, with modifications for satellite specific operations in \highlight{blue}. \textit{FedBuff} specific changes are in \highlightgreen{purple}.}
\newline
Buffer size $D$, $K$ satellites indexed by $k$, $C$ max clients per round, $B$ local minibatch size, $E$ number of local epochs, $\mu$ learning rate, $P_k$ data onboard client $k$, and ground network $G$ indexed by $g$.}\label{alg:fedBuffSat}
\begin{algorithmic}
\algrenewcommand\algorithmicrequire{\textbf{Main Server executes:}}
\algrenewcommand\algorithmicensure{\textbf{ClientUpdate($k$,$w$):}}
\Require 
    \State initialize $w_0$ (global model weights)
\While{not converged, each round $t = 0,1,2,...$} 
    \State{\highlightgreen{Set buffer size} $D \leftarrow$ \highlight{min$(C, K)$}} \Comment{\textit{\highlightgreen{Buffered Update, for secure aggregation}}}
    \State{$S_t \leftarrow $(\highlight{all clients $K$ contact a ground station})} \Comment{\textit{\highlightgreen{Async}}}
    \State{Run \textbf{ClientUpdate}($k,w_t)$   $\forall k\in K$} \Comment{\textit{\highlightgreen{Async}}}
    \State{Initialize buffer counter $d$}
    \While{$d < D$}
        \If {client $k$ contacts $G$ to update}
            \State{$w^k_{t+1} \leftarrow $(\highlightgreen{update from  \textbf{ClientUpdate} $(k,w_t)$})}
    
            \State{$d = d+1$}
            \Comment{\textit{\highlightgreen{Updating current buffer size}}}
            
        \EndIf
    \EndWhile
    \State{$m_t \leftarrow \Sigma_{k \in S_t} n_k$}
    \State{$w_{t+1} \leftarrow \Sigma_{k \in S_t} \frac{n_k}{m_t} w^k_{t+1}$}
    \Comment{\highlightgreen{Round complete, reset $D$}}
\EndWhile 
\newline

\Ensure \Comment{\textit{Run on client/satellite k}}
\State {\highlight{\textbf{Receive}  $w$ from ground station}}
\State{$\beta \leftarrow$ (Split $P_k$ into batches of size $B$)}
\While{\highlightgreen{no access to ground station}}
\For{ batch $b \in \beta$ }
\State{$w_t = \mu \nabla l(w;b)$}
\State{$w \leftarrow w_t + \frac{\mu}{2} || w_t - w ||^2$} \Comment{\highlightgreen{\textit{Similar to \textit{FedProx}, Proximal Term, limits impact of local updates}}}
\EndFor
\EndWhile
\State{\highlight{return $w$ to nearest ground station}}
\end{algorithmic}
\end{algorithm}
\subsection{Space-ification of \textit{FedBuff}: Asynchronous Aggregation}

Compared to the previous two algorithms, \textit{FedBuff} offers a distinct approach due to its asynchronous methodology. Like \textit{FedProx}, each client may perform a variable number of local training epochs, contributing weights that reflect only partial updates. However, the aggregation mechanism operates via an asynchronous buffer system: the global model is updated only once the buffer $D$ is filled, so all clients may not be able to synchronize to the same global model. As clients reach the server at different times during the round, satellites continue training until their next contact with a ground station, thereby minimizing computational idle time. Depending on the latency in reaching a ground station, some clients may generate updates based on slightly outdated global models. These updates are incorporated only if they satisfy the bounded staleness constraint imposed by the server. The complete protocol is outlined in \Cref{alg:fedBuffSat}.

\begin{figure*}[htpb]
\centering
\begin{subfigure}{.48\textwidth}
  \centering
  \includegraphics[width=\linewidth]{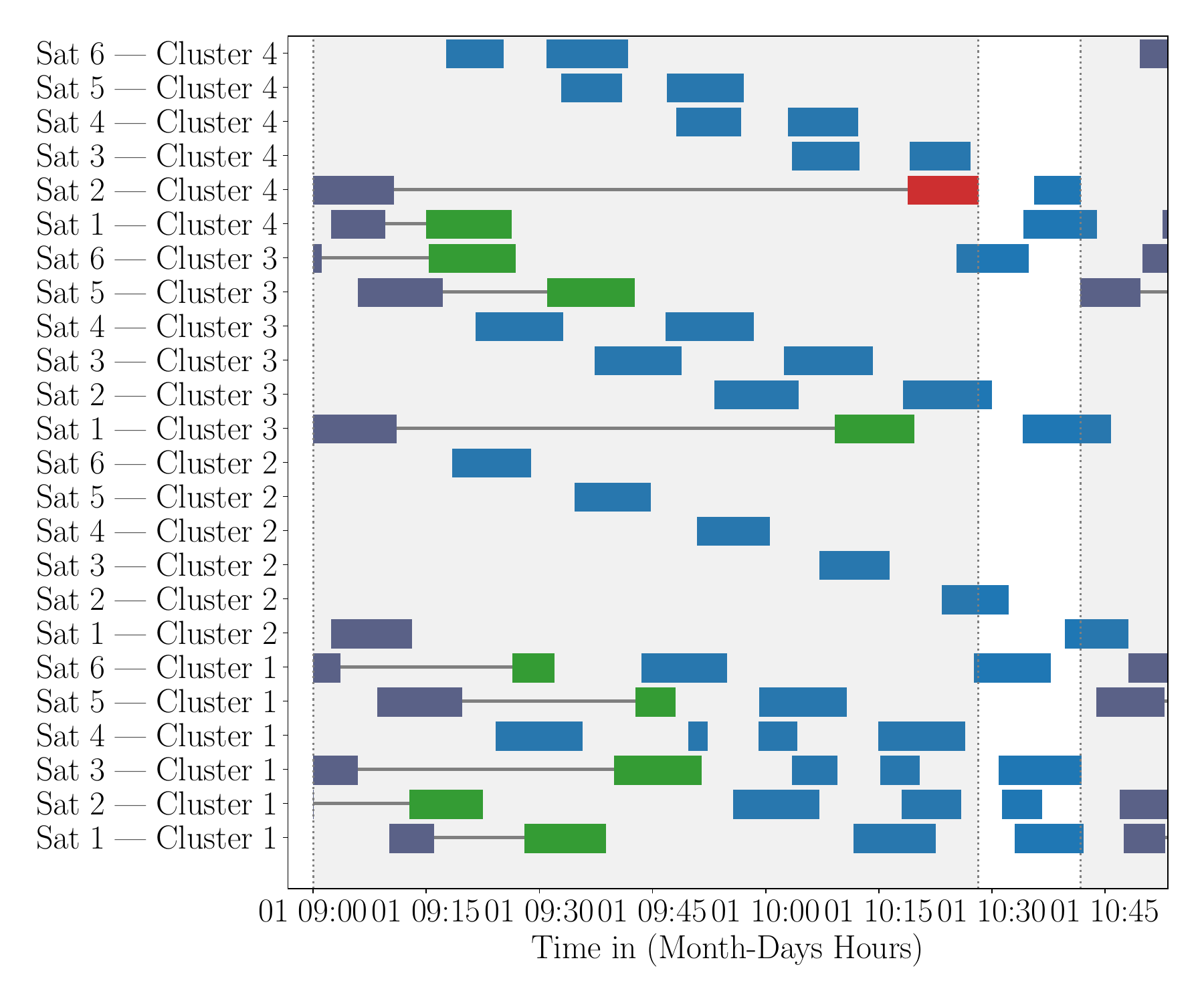}
  \caption{Satellite client access window selection without scheduling. }
  \label{fig:noschedaccess}
\end{subfigure}%
\begin{subfigure}{.48\textwidth}
  \centering
  \includegraphics[width=\linewidth]{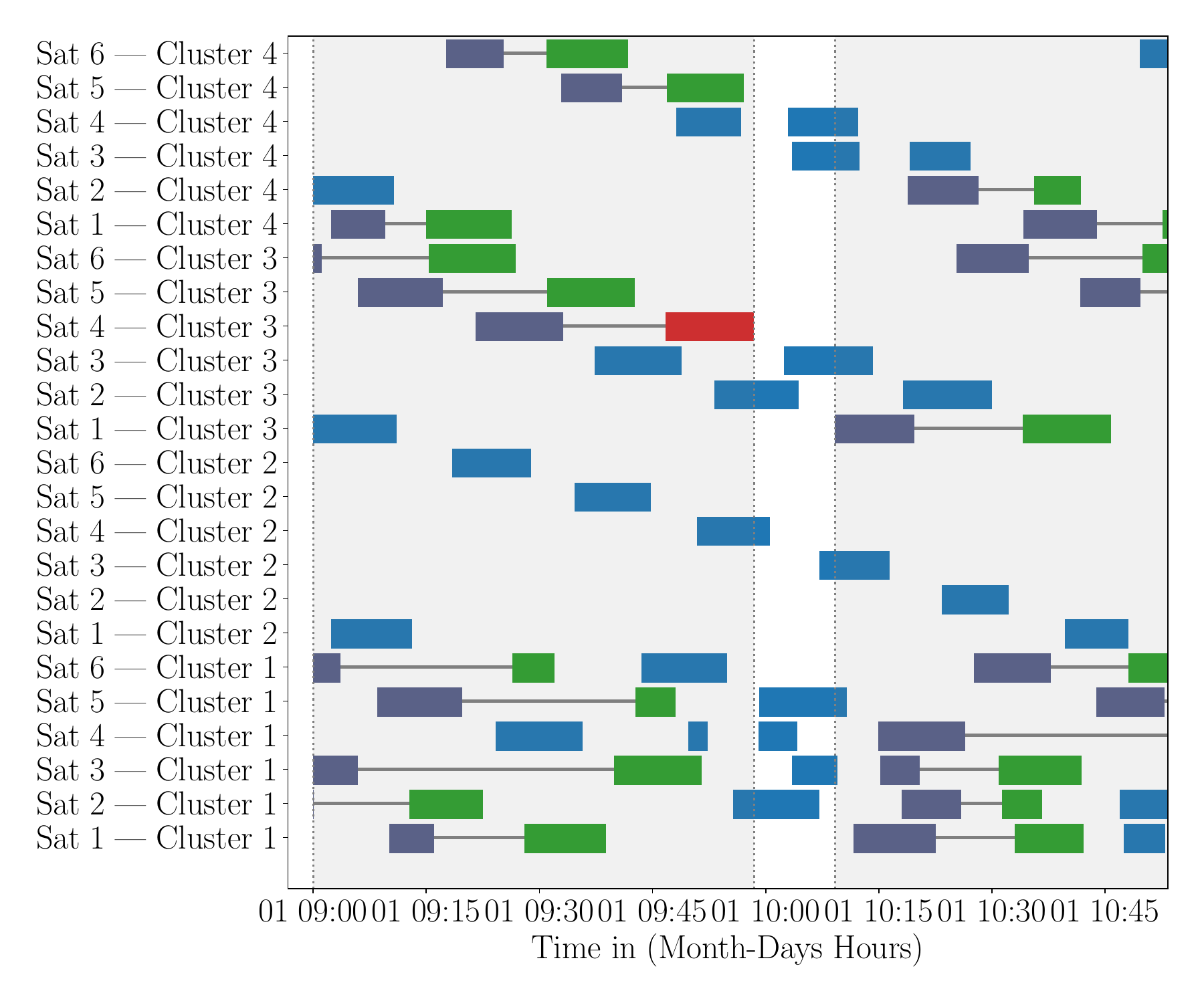}
  \caption{Satellite client access window selection with scheduling. }
  \label{fig:schedaccess}
\end{subfigure}
\caption{Scheduling client selection in FL rounds reduces total aggregation time by prioritizing satellites with shorter combined initial contact and revisit times. Satellites that make initial contact later can still be selected if their revisit times are faster, reducing idle time by nearly half compared to selecting the first satellites to contact a ground station. The figures above show simulations of a Walker-Star constellation with 4 clusters of 6 satellites each and access to 3 ground stations.}
\label{fig:access_small}
\end{figure*}
\section{Satellite-specific FL Augmentations}
\label{sec:augmentations}

After applying the core space-ification process to adapt FL algorithms for satellite environments, further performance gains and efficiency improvements can be realized through modular augmentations. These augmentations, client scheduling and intra-cluster communication (ICC), are designed as enhancements that can be seamlessly integrated with any space-ified FL algorithm to better exploit the predictable orbital dynamics and communication constraints of satellite constellations. Both draw inspiration from prior work \cite{elmahallawy_fedhap_2022, Razmi2022FedSat, Zhai2024FedLEO} but are generalized here for wide compatibility and ready integration with different protocols.

\begin{algorithm}[!htbp]
\caption{\textit{\highlight{FLSchedule}. Once again, the base algorithm in black is referenced from the original \textit{FedAvg} paper \cite{mcmahan_communication-efficient_2016}, with modifications for satellite-specific operations in \highlight{blue}.} \newline 
$K$ satellites indexed by $k$, $C$ max clients per round, ground network $G$ indexed by $g$.}\label{alg:fedAvg2Sat}
\begin{algorithmic}
\algrenewcommand\algorithmicrequire{\textbf{Main server executes:}}
\algrenewcommand\algorithmicensure{\textbf{ClientUpdate($k$,$w$):}}
\algnewcommand\Scheduler{\item[\highlight{\textbf{Scheduler}$(K,G, C, t)$}]}
\Require 
\State{\textit{**Use initialization protocols from selected FL alg}}
\For{each round $t = 0,1,2,\dots$} 
    \State{$S_t \leftarrow$ \highlight{\textbf{Scheduler}$(K,G,C,t)$}}\Comment{\textit{\highlight{Schedule/select clients}}}
    \State{\textbf{Send} $w_t$ to each $k$ in $S_t$}
    \For{each client $k \in S_t$}
    \State{\textit{**Use aggregation protocols from selected FL alg}}
    \EndFor

\EndFor
\Statex
\Scheduler
\State{Initialize $S_t$}
\Comment{\textit{Client list}}
\State{$T \leftarrow $ Start time $t$}
\Comment{\textit{Initialize start of orbit \textbf{propagation}}}
\While{length$(S_t)$ $<C$}
\State{\highlight{\textbf{Propagate} each satellite's $k \in K$ orbital trajectories}}
\If{$k$ can contact any ground station $g\in G$}
    \State{\textbf{Calculate} $k$ to $g$ access times}
    \If{$k \notin S_t$ and contacts any $g \in G$ again}
        \State{$S_t \leftarrow$ append $k$} \Comment{\highlight{\textit{Select \textbf{fastest-returning} $k$}}}
    \EndIf
\EndIf
\EndWhile
\State {\textbf{Return} $S_t$}
\Statex
\Ensure
\State{\textit{**Use \textbf{ClientUpdate} function from selected FL alg.}}

\end{algorithmic}
\end{algorithm}

\subsection{Optimizing Deterministic Orbits, Access Scheduler}

Satellite orbits are deterministic, enabling scheduling of client selections in each FL round\textemdash an opportunity not typically available in terrestrial FL scenarios. By prioritizing satellites with shorter combined initial contact and revisit times, the duration of each round can be significantly reduced, particularly in scenarios with multiple ground station contact points and larger constellations. The potential gains are substantial: for a LEO satellite at 500 km altitude, access and revisit times to a ground station can range from approximately 30 minutes to over 9 hours. Prioritizing satellites with shorter access intervals optimizes aggregation time considerably. To illustrate this concept, access plots for a sample satellite constellation are shown in \Cref{fig:access_small}.

\Cref{alg:fedAvg2Sat} presents the scheduling protocol, referred to as FLSchedule, which can be applied to any space-ified FL algorithm. The protocol follows two main steps. First, the main server precomputes satellite orbits and their corresponding access windows to all ground stations. Second, client selection prioritizes satellites with the shortest total initial contact plus revisit times to available ground stations.

\subsection{Cluster-Based Communication: Intra-Satellite Links} 
\begin{figure}[htbp]
    \centering
    \includegraphics[width=0.9\linewidth]{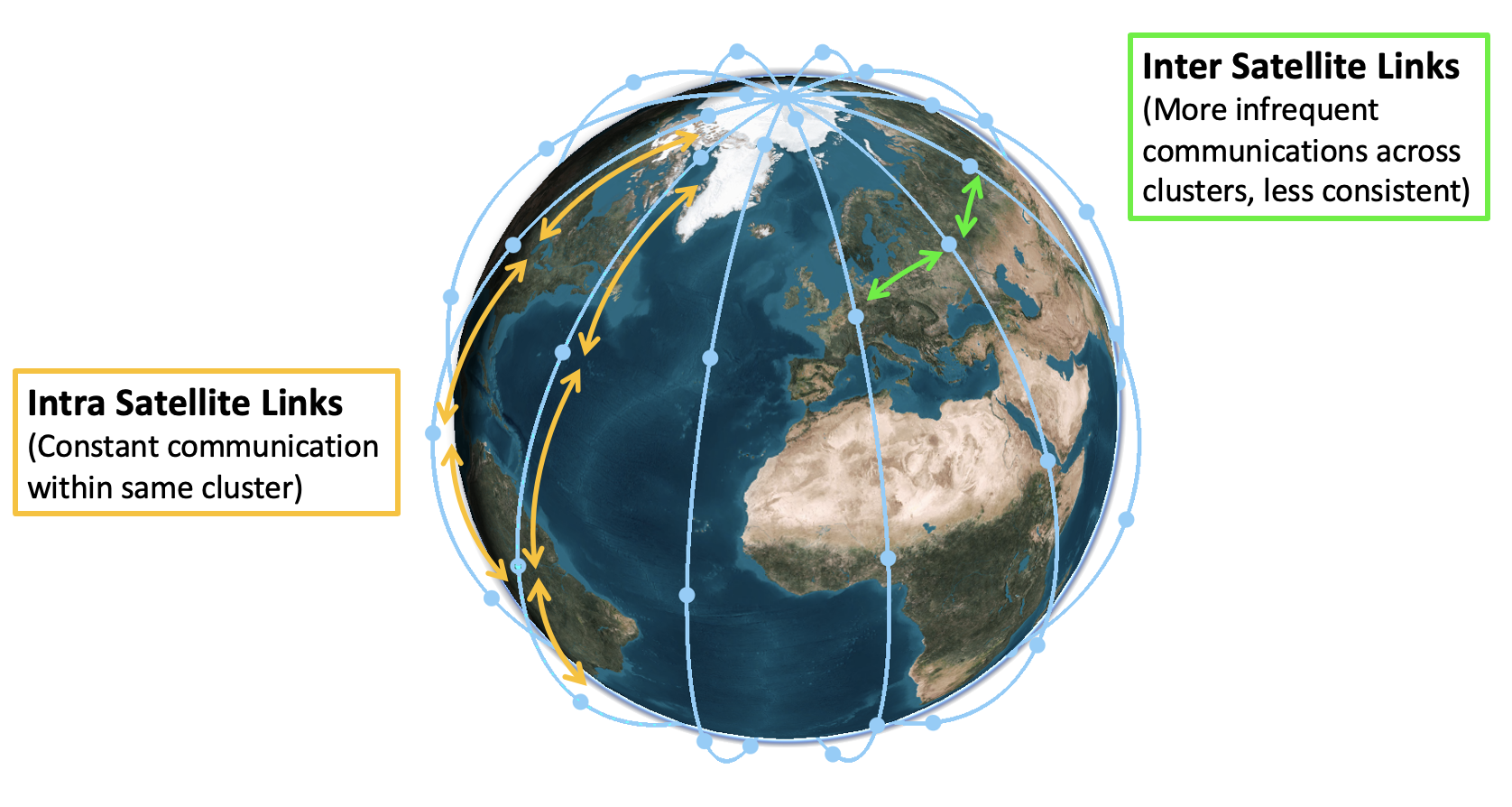}

    \caption{Visual representation of intra- and inter-cluster links (Walker-star topology). Intra-cluster communication occurs between adjacent satellites of the same cluster, and may be persistent. In contrast, inter-cluster communication occurs across neighboring clusters, typically over a finite communication window in which the satellites are within range.}

    \label{fig:ISLs}
\end{figure}

For large constellations, satellites can exchange data through intra- or inter-satellite links. To ensure stable communication, we focus here on intra-cluster communication, i.e., data transfer between satellites within the same orbital cluster, displayed in \Cref{fig:ISLs}. When satellites in a cluster are closely spaced, continuous line-of-sight can be maintained along the orbital plane, allowing data to be relayed through peers instead of requiring direct ground station contact. Proper intra-CC scheduling can significantly reduce idle and revisit times across the constellation, though a minimum cluster size is required for such links, depending on altitude (e.g., about ten satellites at 500 km in LEO).

\Cref{alg:fedAvg3Sat} extends the space-ified FL framework to include intra-CC. In this setting, client selection accounts not only for ground station access windows but also for intra-cluster satellite-to-satellite communication. Additional checks are introduced to ensure that at least one satellite in the cluster, not only the training node, can downlink to a ground station. Finally, when both the original training satellite and a peer cluster member can establish contact, priority is given to the original satellite to return updates, maintaining consistency in the aggregation process.

\begin{algorithm}[!ht]
\caption{\textit{\highlight{FLIntraCC}}. \textit{Once again, the base algorithm in black is referenced from the original \textit{FedAvg} paper \cite{mcmahan_communication-efficient_2016}, with modifications for satellite-specific operations in \highlight{blue}.} $K$ satellites indexed by $k$, $C$ max clients per round, ground network $G$ indexed by $g$.}\label{alg:fedAvg3Sat}
\begin{algorithmic}
\algrenewcommand\algorithmicrequire{\textbf{Main server executes:}}
\algrenewcommand\algorithmicensure{\textbf{ClientUpdate($k$,$w$):}}
\algnewcommand\IntraScheduler{\item[\highlight{\textbf{IntraCC}$(K, G, C, t)$}]}
\Require 
    \State{\textit{**Use initialization protocols from selected FL alg}}
\For{each round $t = 0,1,2,\dots$} 
    \State{$S_t \leftarrow$ \textbf{IntraCC$(K,G,C,t)$} } \Comment{\textit{\highlight{Select clients}}}
    \State{\textbf{Send} $w_t$ to each $k$ in $S_t$}
    \For{each client $k \in S_t$ }
        \State{\textit{**Use aggregation protocols from specific FL alg}}
    \EndFor
\EndFor
\Statex

\IntraScheduler
\State{Initialize $S_t$} \Comment{\textit{Client list}}
\State{\highlight{Initialize each cluster $S_{p}$}} \Comment{\highlight{\textit{Track each cluster's contacts}}}
\State{$T \leftarrow $ Start time $t$} \Comment{\textit{Initialize start of orbit \textbf{propagation}}}
\While{length($S_t)<C$}
\State{\textbf{Propagate} each satellite's $k \in K$ orbital trajectories}
\If{$k$ can contact any ground station $g \in G$}
    \State{\textbf{Calculate} $k$ to $g$ access times}    
    \State{\highlight{\textbf{Add} $k$ to it's cluster's $S_{passed}$}}
    \If{$k \notin S_t$ and contacts any $g \in G$ again}
        \State{$S_t \leftarrow$ append $k$} \Comment{\textit{Select \textbf{fastest}-returning $k$}}
        \If{length($S_t) = C$}
            \State{\textbf{Return} $S_t$}
        \EndIf
    \EndIf
    \If{\highlight{fellow cluster satellite $k_i \in S_{p}$ and $k_i \notin S_t$}}
        \State{$S_t \leftarrow$ append $k$} \Comment{\highlight{\textit{Select \textbf{IntraCC} access}}}
        \If{length($S_t) = C$}
            \State{\textbf{Return} $S_t$}
        \EndIf   
    \EndIf
\EndIf
\EndWhile
\State{\textbf{Return} $S_t$}
\Statex

\Ensure 
\State{\textit{**Use \textbf{ClientUpdate} function from selected FL alg.}}

\end{algorithmic}
\end{algorithm}

\section{Experimental Setup}\label{sec:design_platforms_sims_and_hardware}

We rigorously test the feasibility of applying our space-ified terrestrial FL algorithms in the orbital environment by evaluating how constellation size, cluster configuration, and ground station availability affect training efficiency and model performance. To capture this at scale, we conduct an extensive set of simulations, 768 distinct satellite and network scenarios, covering a wide range of operational conditions. This breadth enables us to assess not only how well space-ified versions of \textit{FedAvg}, \textit{FedProx}, and \textit{FedBuff} perform, but also whether such approaches are practical and worthwhile for real satellite missions.

To perform these evaluations, we integrate two domain-specific simulation tools and FL frameworks. Communication links, both satellite-to-satellite and satellite-to-ground, are simulated using the Systems Tool Kit (STK) analysis software. STK is a physics-based modeling environment widely used in aerospace and industrial applications to analyze air, naval, and space platforms in realistic mission contexts \cite{rainey_space_2024}. Satellite constellations are modeled with precise orbital parameters, and access windows between satellites and ground stations are generated. These windows are exported as CSV files to serve as input for \textit{Flower}, an open-source FL framework \cite{beutel_flower_2020}. \textit{Flower} allows us to simulate federated ML clients and modify the FL strategies used for organizing satellite clients. Using the STK-generated communication windows, appropriate client selection protocols are applied to execute the desired FL algorithms at each point in time.

We evaluate our proposed space-ified FL algorithms over a wide range of synthetic satellite and network configurations, testing 768 scenarios outlined in \Cref{tab:clflConfigs}. This section details the satellite constellation and ground network setups, datasets, simulation workflow, and hardware assumptions. Each experiment is repeated with five random seeds, and results are averaged across trials. Simulations are conducted over a three-month orbital scenario from April 14, 2024, to July 13, 2024. FL algorithms are limited to a maximum of 500 training rounds or the maximum number of rounds possible within the simulation timeframe, whichever is first.

\begin{table}[htbp]
    \renewcommand{\arraystretch}{1.0}
    \centering
    \caption{Parameter configurations tested across space-ified FL algorithms (\textit{FedAvg}, \textit{FedProx}, \textit{FedBuff}). Sweeps cover clusters, satellites per cluster, and ground stations, evaluated on FEMNIST, leading to 768 total tested configurations.}
    \label{tab:clflConfigs}
    \scriptsize 
    \setlength{\tabcolsep}{4pt} 
    \begin{tabular}{l l c c c}
    \toprule
    \textbf{FL Alg} & \textbf{Extension} & \textbf{Clusters} & \textbf{Sats/Cluster} & \textbf{Ground Stations} \\ 
    \midrule
    FedAvg  & Base      & $1,2,5,10$ & $1,2,5,10$ & $1,2,3,5,10,13$ \\
            & Schedule  & $1,2,5,10$ & $1,2,5,10$ & $1,2,3,5,10,13$ \\
            & IntraCC   & $1,2,5,10$ & $1,2,5,10$ & $1,2,3,5,10,13$ \\
    \midrule
    FedProx & Base      & $1,2,5,10$ & $1,2,5,10$ & $1,2,3,5,10,13$ \\
            & Schedule  & $1,2,5,10$ & $1,2,5,10$ & $1,2,3,5,10,13$ \\
            & Sched V2  & $1,2,5,10$ & $1,2,5,10$ & $1,2,3,5,10,13$ \\
            & IntraCC   & $1,2,5,10$ & $1,2,5,10$ & $1,2,3,5,10,13$ \\
    \midrule
    FedBuff & Base      & $1,2,5,10$ & $1,2,5,10$ & $1,2,3,5,10,13$ \\
    \bottomrule
    \end{tabular}
    
\end{table}

\subsection{Satellite and Ground Station Configuration Space}

We analyze three key satellite configuration parameters: the number of satellite clusters, the number of satellites per cluster, and the number of ground stations available for communication. Sweeping these parameters allows us to test several important operational scenarios, including: (1) ground-station-only communications, (2) scheduling-based FL operations, and (3) intra-cluster communications.

\textbf{Constellation Parameters.} We model synthetic Walker-Star constellations with up to 100 satellites organized into clusters containing up to 10 satellites each. The number of clusters is varied across $\{1,2,5,10\}$, and the number of satellites per cluster is varied across $\{1,2,5,10\}$. Satellites in each cluster share an orbital plane, with uniform RAAN spacing across clusters and uniform true anomaly spacing within clusters. Inspired by sun-synchronous LEO Earth observation constellations, all satellites are in circular, polar orbits (eccentricity $=0$, inclination $=90^\circ$) at an altitude of 500 km above the Earth's surface.

\begin{table}[htbp]
\centering
\caption{Walker-Star Constellation Orbital Parameters}
\label{tab:orbital_params}
\begin{tabular}{c c}
\toprule
\textbf{Parameter} & \textbf{Value} \\
\midrule
Orbit type & Circular, Walker-Star \\
Altitude & 500 km \\
Inclination & 90$^\circ$ (polar) \\
Eccentricity & 0 \\
RAAN spacing & Uniform by cluster \\
Argument of perigee & 0$^\circ$ \\
True anomaly spacing & Uniform within cluster \\
\bottomrule
\end{tabular}
\end{table}

\begin{figure}[htb]
    \centering
    \includegraphics[width=0.95\linewidth]{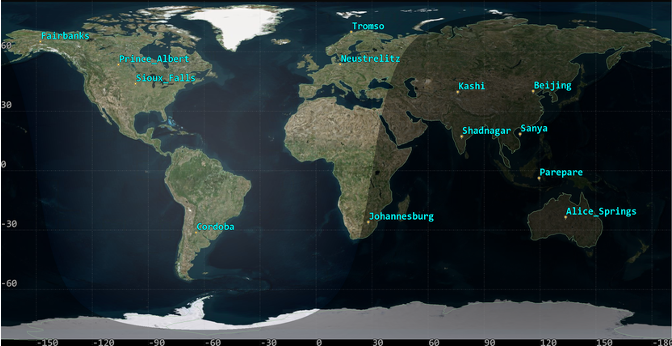}
    \caption{Ground station facilities used for parameter sweep simulations, with 13 communication locations inspired from the International Ground Station (IGS) network \cite{wulder_global_2016}. The different locations are set to be in: Sioux Falls (US), Sanya (China), Johannesburg (South Africa), Cordoba (Argentina), Tromso (Norway), Kashi (China), Beijing (China), Neustrelitz (Germany), Parepare (Indonesia), Alice Springs (Australia), Fairbanks (US), Prince Albert (Canada), and Shadnagar (India).}
    \label{fig:groundstations}
\end{figure}

\textbf{Ground Station Placements.} Motivated by the International Ground Station (IGS) Network \cite{wulder_global_2016}, which has supported the operation of the Landsat 8 and 9 Earth observation missions, we adopt a set of 13 ground station sites distributed globally, as shown in \Cref{fig:groundstations}. This network was designed to provide broad geographic distribution to maximize contact opportunities across the globe for satellite communication. For experimental evaluations, subsets of the network are selected with sizes drawn from $\{1,2,3,5,10,13\}$. The 13-station configuration corresponds to the full IGS-inspired network, while smaller networks represent reduced-coverage cases. The specific sites in each configuration are detailed in \Cref{tab:groundStationNamesCoords}.

\begin{table}[htbp]
\centering
\caption{Ground station names and latitude/longitude coordinates used in different experimental sweeps. Each row shows one station, with the left column spanning the total number of stations for that configuration.}
\footnotesize
\setlength{\tabcolsep}{4pt}
\renewcommand{\arraystretch}{1.2}
\resizebox{0.95\linewidth}{!}{%
\begin{tabular}{c l l}
\toprule
\textbf{\# Ground Stations} & \textbf{Names} & \textbf{Coordinates (Lat, Lon)} \\ \midrule
\multirow{1}{*}{1}  & Sioux Falls & (43.55, -96.72) \\ \hline
\multirow{1}{*}{2}  & Sanya & (18.25, 109.5) \\ \hline
\multirow{1}{*}{3}  & Johannesburg & (-26.2, 28.03) \\ \hline
\multirow{2}{*}{5}  & Cordoba & (-31.4, -64.18) \\ 
                     & Tromso & (69.65, 18.95) \\ \hline
\multirow{5}{*}{10} & Kashi & (39.1, 77.2) \\
                     & Beijing & (39.9, 116.4) \\
                     & Neustrelitz & (53.1, 13.1) \\
                     & Parepare & (-2.99, 119.8) \\
                     & Alice Springs & (-25.1, 133.9) \\ \hline
\multirow{3}{*}{13} & Fairbanks & (64.8, -147.7) \\
                     & Prince Albert & (53.2, -105.7) \\
                     & Shadnagar & (17.4, 78.5) \\
\bottomrule
\end{tabular}
}
\label{tab:groundStationNamesCoords}
\end{table}

\subsection{FEMNIST dataset}

To evaluate performance under heterogeneous data distributions, we use the FEMNIST dataset \cite{caldas_leaf_2018}. FEMNIST partitions the Extended MNIST dataset \cite{cohen_emnist_2017} such that each client receives data specific to a writer. While handwritten characters are not directly representative of satellite imagery, the non-IID heterogeneity mimics realistic satellite data collection: images vary slightly due to orbital timing, lighting, and weather conditions, even for satellites in the same cluster.

Each client model in our experiments has 47k parameters and is trained on 200--350 samples per satellite. We assume a standard CubeSat onboard computer, e.g., the SpaceCloud iX5-106, capable of 40 GFlops/s \cite{flordal2021spacecloud}. Each training epoch requires approximately 98 MFlops, while sending the 47k parameter (186 KB) model between satellites or to ground stations is feasible over telemetry systems, such as Planet Lab's Doves, which communicate at 580 Mbps \cite{PlanetLabs_PSSceneImagerySpec_2023}.

\section{Results}\label{sec:experiments_results}

Our experiments reveal that, with carefully designed satellite constellations and scheduling strategies, space-ified FL algorithms with proper performance augmentations can achieve high accuracy while completing training in realistic orbital timeframes, reducing training durations from months to days. This demonstrates that FL in space is not only feasible but can be made efficient through the right combination of constellation size, intra-cluster communication, and ground station coverage.

We highlight three key takeaways from our simulation sweeps. First, frequent and well-placed access points are critical. Larger ground station networks and intra-satellite links drastically reduce FL round durations, enabling faster convergence. Second, cluster composition outperforms constellation size. Increasing the number of satellites per cluster reduces idle time and accelerates aggregation more effectively than simply adding clusters. Third, scheduling must balance speed and training completeness. Aggressive scheduling can reduce total round durations but may harm accuracy unless a minimum number of local epochs per satellite is ensured.

Below, we explore these findings in depth through three complementary analyses: (1) model accuracy, (2) FL round durations, and (3) satellite idle time. Together, these metrics quantify both the performance and efficiency of FL algorithms under realistic orbital constraints and highlight practical considerations for designing space-based FL systems.

\begin{figure*}[ht]
\begin{minipage}{\textwidth}
    \begin{subfigure}{\textwidth}
    \centering
    \includegraphics[trim ={0 0 1cm 0},clip,width=\textwidth]{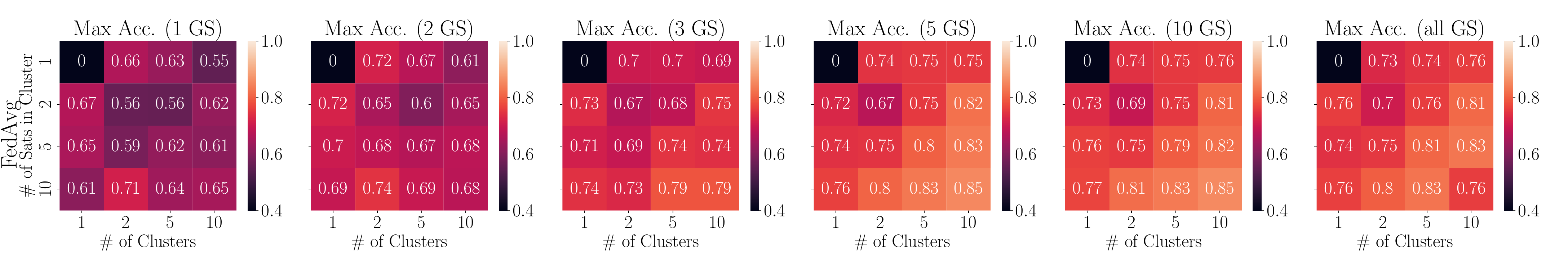}
    \caption{FedAvg with space-ification, accuracy heatmaps}
    \label{fedavg:acc}
    \end{subfigure}

    \begin{subfigure}{\textwidth}
    \centering
    \includegraphics[trim ={0 0 1cm 0},clip,width=\textwidth]{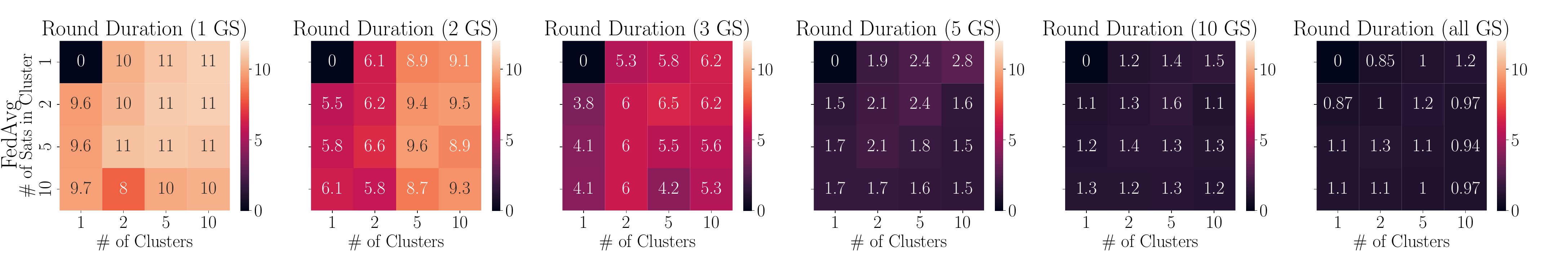}
    \caption{FedAvg with space-ification, round duration heatmaps}
    \label{fedavg10:duration}
    \end{subfigure}

    \begin{subfigure}{\textwidth}
    \centering
    \includegraphics[trim ={0 0 1cm 0},clip,width=\textwidth]{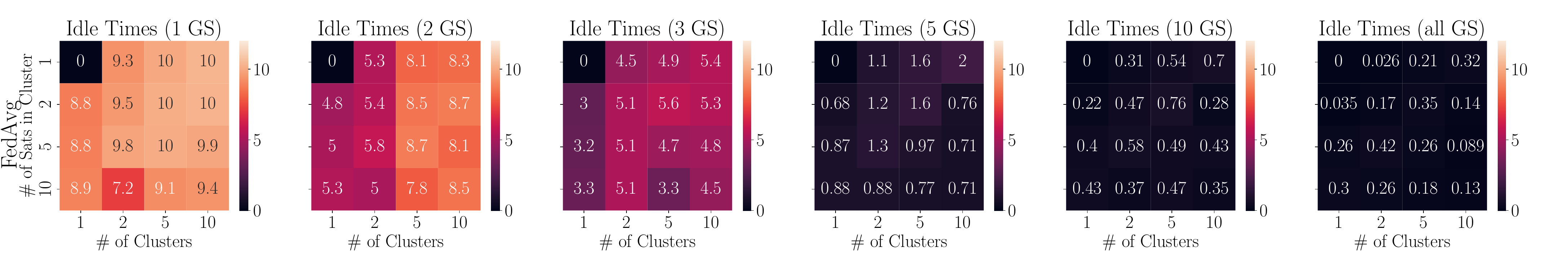}
    \caption{FedAvg with space-ification, idle time heatmaps}
    \label{fedavg10:iavg}
    \end{subfigure}

    \caption{An example set of heatmaps for FedAvg with space-ification, outlining performances in accuracy, FL round durations, and idle time. Even with \textit{FedAvg}, we see that with large enough constellation sizes and ground station networks, enough opportunities for access windows can be made to reach convergence. Proper comparisons against performance of other algorithms can be found in the larger span of heatmaps in \Cref{fig:accs,fig:base_durations,fig:idleavg_normal}, highlighting the importance of the augmentations on FL algorithm performance. }\label{fig:fedavgheatmap}
\end{minipage}
\end{figure*}
\subsection{Reading Heatmaps}
\label{app:readingheatmaps}

\begin{figure*}[ht]
\begin{minipage}{\textwidth}

    \begin{subfigure}{\textwidth}
    \centering
    \includegraphics[trim={0 1cm 0 0.95cm},clip,width=\textwidth]{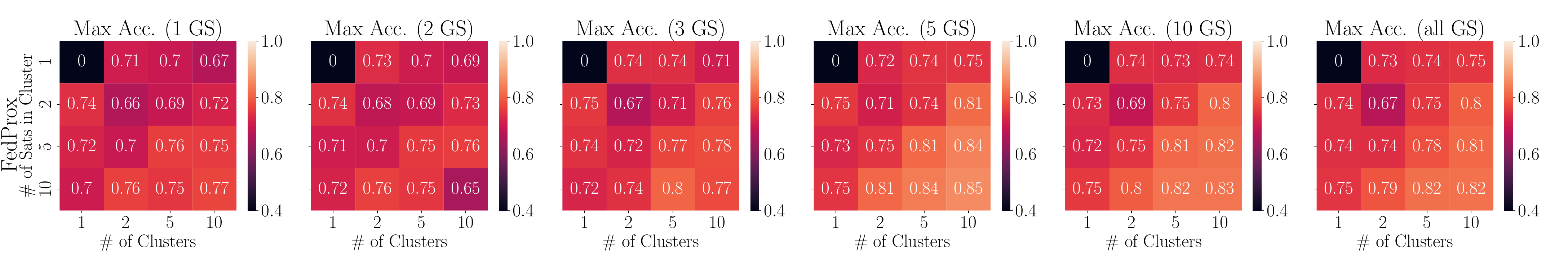}
    \label{fedprox:acc}
    \end{subfigure}
    
    \begin{subfigure}{\textwidth}
    \centering
    \includegraphics[trim={0 1cm 0 0.95cm},clip,width=\textwidth]{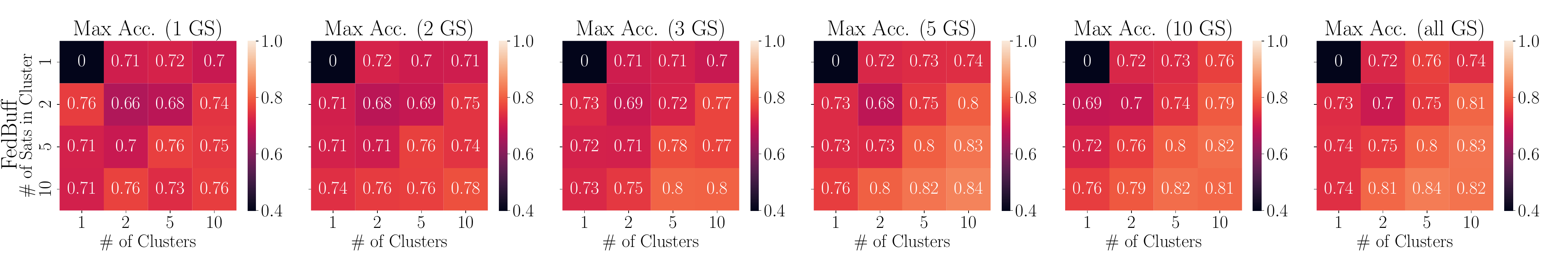}
    \caption{Baseline FL Algorithms, with no Scheduling or Intra Satellite Communications}
    
    \vspace{0.5em}
    \label{fedBuff:acc}
    \end{subfigure}
    
    \begin{subfigure}{\textwidth}
    \centering
    \includegraphics[trim={0 1cm 0 0.95cm},clip,width=\textwidth]{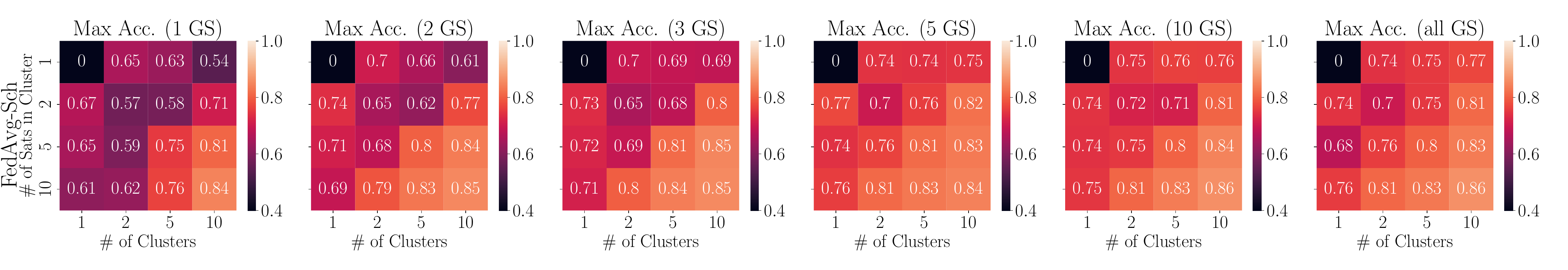}
    \label{fedavg2:acc}
    \end{subfigure}

    \begin{subfigure}{\textwidth}
    \centering
    \includegraphics[trim={0 1cm 0 0.95cm},clip,width=\textwidth]{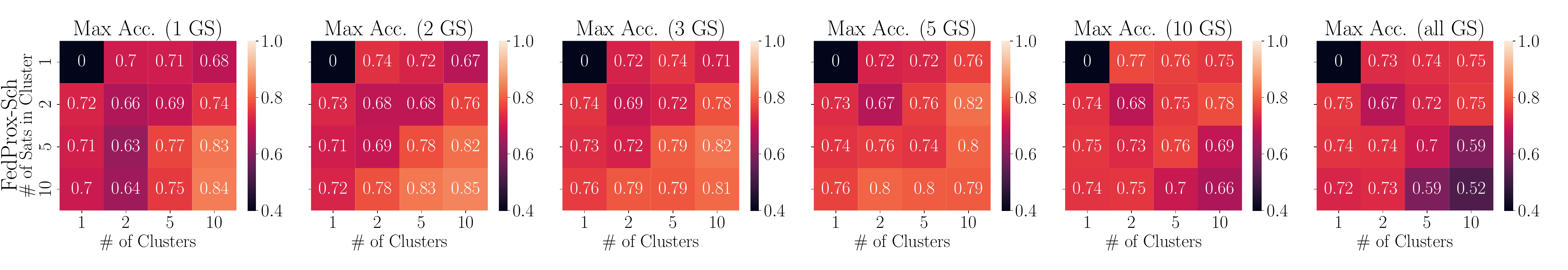}
    \label{fedprox2:acc}
    \end{subfigure}
    
    \begin{subfigure}{\textwidth}
    \centering
    \includegraphics[trim={0 1cm 0 0.95cm},clip,width=\textwidth]{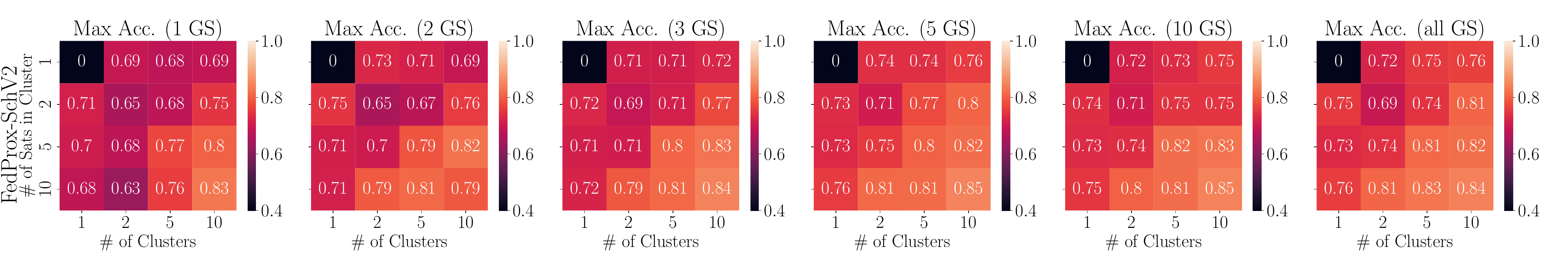}
    \caption{FL Algorithms with a Scheduler, but without Intra Satellite Communications}
    
    \vspace{0.5em}
    \label{fedprox22:acc}
    \end{subfigure}

    \begin{subfigure}{\textwidth}
    \centering
    \includegraphics[trim={0cm 0 0cm 0},clip,width=\textwidth]{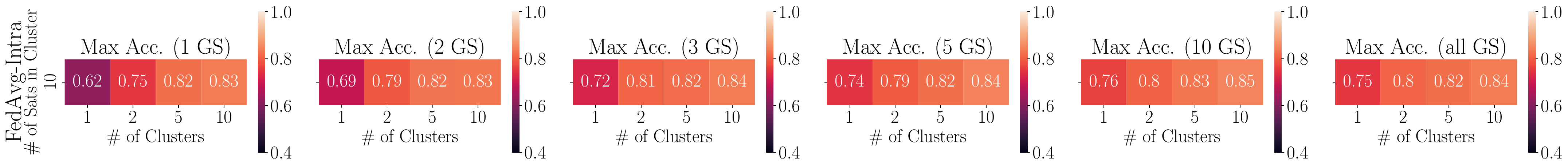}
    \label{fedavg3:acc}
    \end{subfigure}

    \begin{subfigure}{\textwidth}
    \centering
    \includegraphics[trim={0cm 0 0cm 0},clip,width=\textwidth]{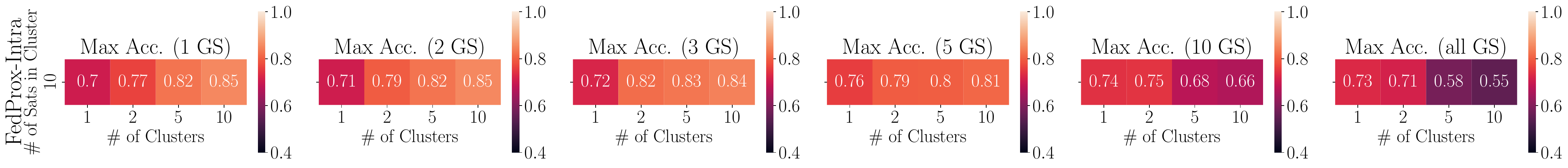}
    \caption{FL Algorithms with both a Scheduler, and enabled Intra Satellite Communications}
    \label{fedprox3:acc}
    \end{subfigure}

    \caption{Heatmaps depicting the maximum accuracy reached in the training of each satellite simulations, testing configuration parameters of varying \#s of clusters, \#s of satellites per cluster, and \#s of ground stations available to connect in the network. Multiple FL algorithms were implemented and tested, specifically \textit{FedAvg}, \textit{FedProx}, and \textit{FedBuff}, and measured against versions of the same algorithm but with scheduled and Intra SL enabled communications. All algorithms if provided enough aggregation opportunities could reach more than 80\% of accuracy. This was typically possible by optimizing for existing access windows through scheduling, or through the addition of more communication points through larger ground station networks or ISL enabled communications.}\label{fig:accs}
\end{minipage}
\end{figure*}

\begin{figure*}[ht]

    \begin{subfigure}{\textwidth}
    \centering
    \includegraphics[trim={0 1cm 0 0},clip,width=\textwidth]{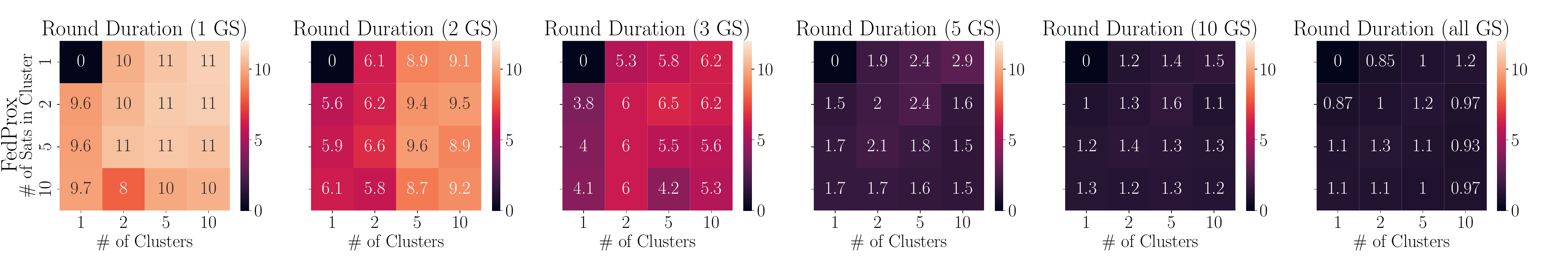}
    \label{fedProx:duration}
    \end{subfigure}

    \begin{subfigure}{\textwidth}
    \centering
    \includegraphics[trim={0 1cm 0 0},clip,width=\textwidth]{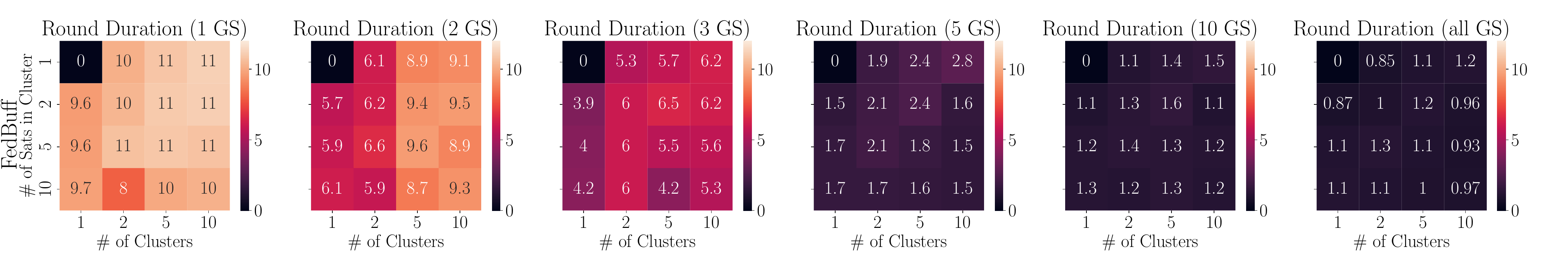}
    \caption{Baseline FL Algorithms, with no Scheduling or Intra Satellite Communications}
    \vspace{0.5em}
    \label{baseline:duration}
    \end{subfigure}

    \begin{subfigure}{\textwidth}
    \centering
    \includegraphics[trim={0 1cm 0 0},clip,width=\textwidth]{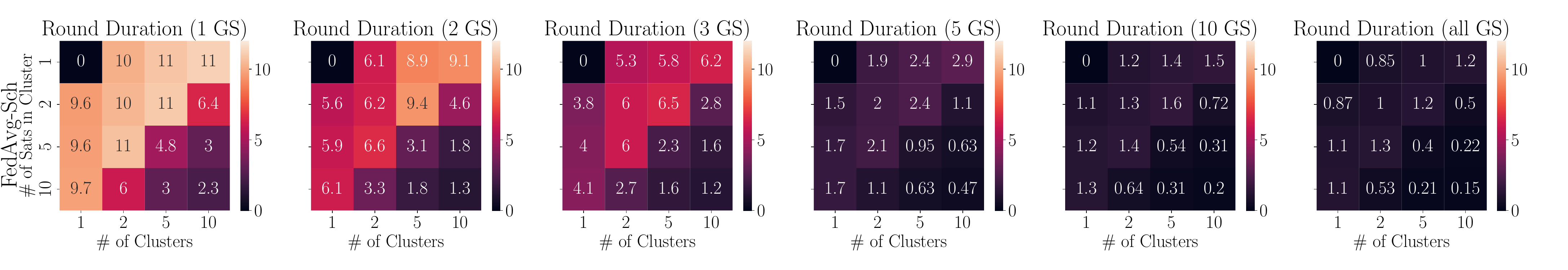}
    \label{fedavg2_10:duration}
    \end{subfigure}

    \begin{subfigure}{\textwidth}
    \centering
    \includegraphics[trim={0 1cm 0 0},clip,width=\textwidth]{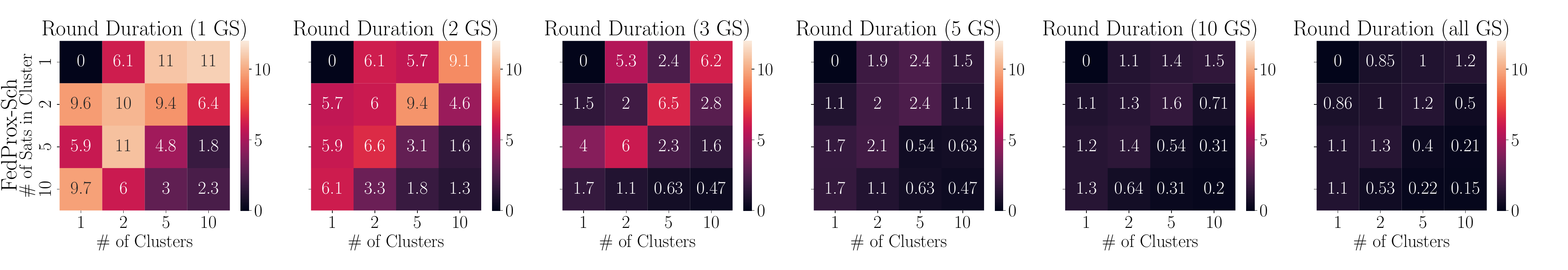}
    \label{fedProx2:duration}
    \end{subfigure}

    \begin{subfigure}{\textwidth}
    \centering
    \includegraphics[trim={0 1cm 0 0},clip,width=\textwidth]{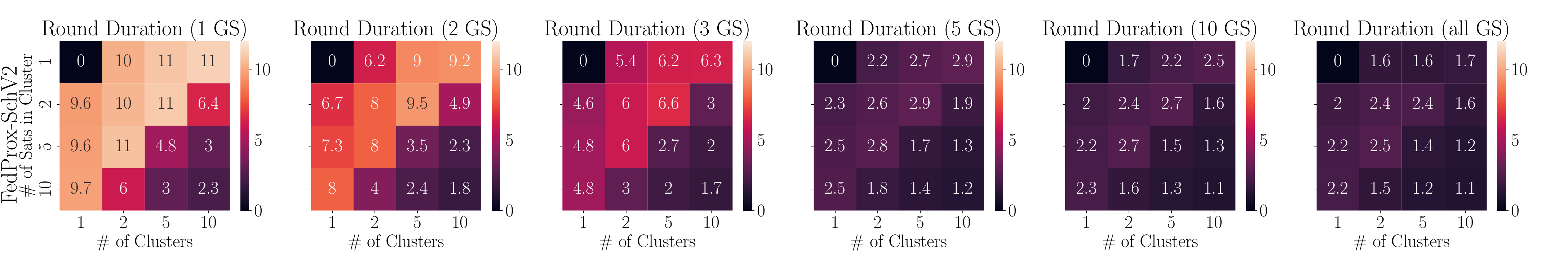}
    \caption{FL Algorithms with a Scheduler, but without Intra Satellite Communications}
    \label{fedBuff2:duration}
    \end{subfigure}
    
    \begin{subfigure}{\textwidth}
    \centering
    \includegraphics[trim={0 0cm 0 0cm},clip,width=\textwidth]{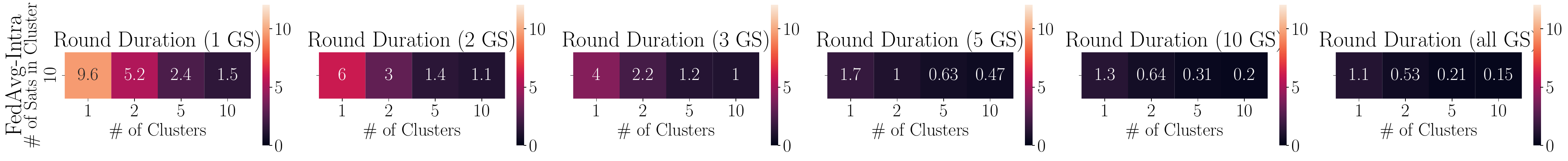}
    \label{fedavg3:dur}
    \end{subfigure}

    \begin{subfigure}{\textwidth}
    \centering
    \includegraphics[trim={0 0cm 0 0cm},clip,width=\textwidth]{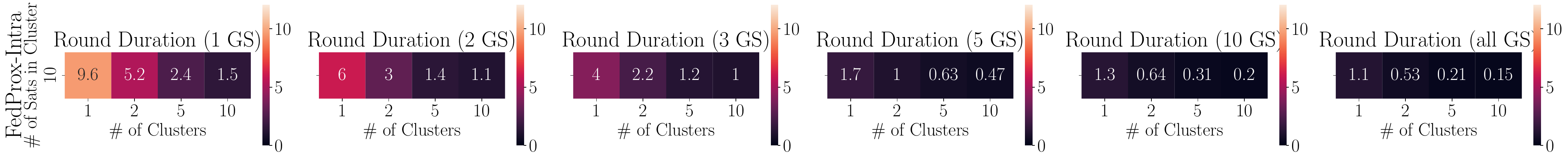}
    \caption{FL Algorithms with both a Scheduler, and enabled Intra Satellite Communications}
    \label{fedprox3:dur}
    
    \end{subfigure}
    \caption{Heatmaps depicting the average round duration per hour taken for FL training and aggregation, again testing changes in the \#s of clusters, \#s of satellites per cluster, and \#s of ground stations available to connect in the network. \textit{FedAvg}, \textit{FedProx}, and \textit{FedBuff}, are tested measured against versions of the same algorithm but with scheduled and Intra SL enabled communications. The number of ground stations seems to have the largest impact on round duration length, with clear drops in duration when testing 1, 2, 3, and 5 ground stations. However, a plateauing effect is seen with any additional ground stations added, suggesting a optimal ratio of satellite to ground stations in a constellation network.}\label{fig:base_durations}
\end{figure*}

To interpret our results, we summarize the performance across the various constellation configurations and FL algorithms using heatmaps, such as those in \Cref{fig:accs,fig:base_durations,fig:idleavg_normal}. Each square represents a specific constellation–algorithm combination, with the top-left square set to 0, as a single satellite cannot perform FL. Rows indicate satellites per cluster {$1,2,5,10$}, columns indicate number of clusters {$1,2,5,10$}, and the total satellites per experiment is derived by multiplying these values. Columns across heatmaps also encode the number of ground stations {$1,2,3,5,10,13$}.  

We evaluated several space-ified FL algorithms (\textit{FedAvg}, \textit{FedProx}, \textit{FedBuff}, \textit{FedAvgSch}, \textit{FedAvgSch2}, \textit{FedProxSch}, \textit{FedAvgIntraSL}, \textit{FedProxIntraSL}) under our three metrics of accuracy, FL round duration, and satellite idle time. An abbreviated example of these heatmaps for the base space-ified algorithm \textit{FedAvg} is displayed in \Cref{fig:fedavgheatmap}.

\subsection{Accuracy: A Baseline Comparison of Performance}
\label{subsec:accuracy}

We first evaluate the baseline accuracy of our space-ified FL algorithms before considering computation or time-based optimizations. Accuracy results, measured on the FEMNIST dataset, are summarized in the heatmaps in \Cref{fig:accs}, which compare the base algorithms \textit{FedAvg}, \textit{FedProx}, and \textit{FedBuff} to their counterparts incorporating scheduling and intra-cluster communications.

\begin{figure*}[htbp]
    \centering
    \includegraphics[width=\linewidth]{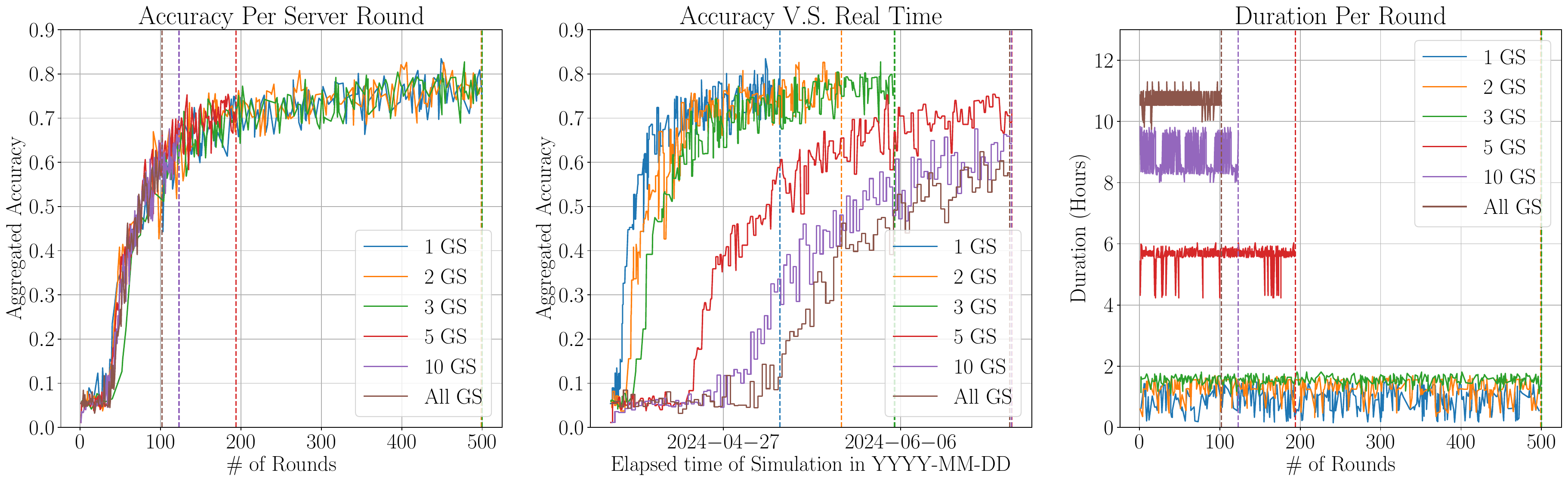}
    \caption{Performance of \textit{FedAvg} with 5 clusters and 10 satellites per cluster. (Left) Accuracy versus server rounds shows smaller networks with 1 or 2 ground stations struggle to perform as many aggregation rounds. (Center) Accuracy versus simulation time highlights slow convergence for small ground station networks even after 3 months. (Right) Round duration plot reveals large increases in FL round time based on the number of available ground stations.}
    \label{fig:acc_line_fedAvg}
\end{figure*}

\begin{figure*}[htp]
    \centering
    \includegraphics[width=\textwidth]{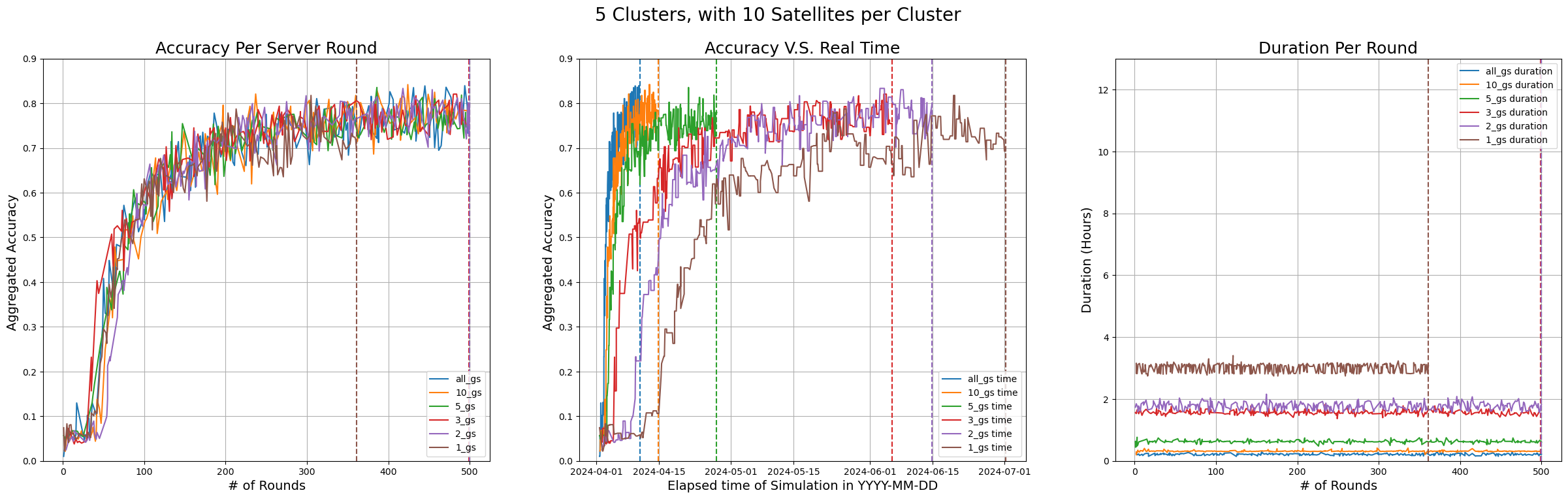}

    \caption{Performance of \textit{FedAvgSch} with 5 clusters and 10 satellites per cluster. (Left) Accuracy vs. server rounds demonstrates most ground station networks reaching close to 500 rounds within 3 months. (Center) Accuracy vs. simulation time illustrates scheduling benefits, with simulation times dropping to 10 days for the largest network. (Right) Round duration plots show reduced variance through scheduling.}
    \label{fig:acc_line_fedAvg2}
\end{figure*}

Even working under extremely restrictive satellite access communication constraints, all FL algorithms were able to reach at least an 80\% accuracy rate if provided \textit{frequent enough access points} to perform aggregation. Constellations with limited ground station coverage exhibited lower accuracy, as indicated by the darker gradients in the leftmost heatmaps. Accuracy improved with larger ground station networks or additional intra-cluster links, highlighting the importance of access frequency. Notably, \textit{FedAvg}, often considered suboptimal in prior works, achieves competitive performance under frequent aggregation conditions.

However, if each client is allotted too short a training period, overall performance was found to decrease. Overly aggressive optimization of initial visit and revisit windows prevented satellites from performing sufficient computation per round. This trend was observed for \textit{FedProx} when applying the scheduler to create \textit{FedProxSch}; without scheduling, \textit{FedProx} performed better in scenarios with many ground stations. This behavior likely results from the absence of a minimum-epoch requirement per client, limiting model updates per round. To address this, we introduced in \textit{FedProxSchV2} a minimum number of local epochs enforcement before returning parameters, mitigating accuracy loss across larger constellations.  

Conversely, \textit{FedBuff} reached high performance through the benefits of asynchronous updates, as satellites were allowed to continue local training until an aggregation opportunity occurred. Compared to other methods that were forced to stop training to ensure synchronous aggregation across all clients, \textit{FedBuff} would still take in client updates from any participant as long as the staleness was higher than a certain threshold. This mechanism allowed \textit{FedBuff} to maintain high accuracy even for smaller constellations, when synchronous methods struggled to converge.

Achieving these accuracy levels, however, requires trade-offs in the time needed for each aggregation round. As noted in \Cref{sec:design_platforms_sims_and_hardware}, experiments were terminated either after 500 aggregation rounds or three months of orbital simulation. This can be a deterring factor for satellite operators, as algorithms that do not guarantee both performance and speed may be impractical. Without sufficient scheduling or ground station coverage, smaller constellations could not surpass 60\% accuracy within the three-month period. 

Figures \ref{fig:acc_line_fedAvg} and \ref{fig:acc_line_fedAvg2} illustrate this effect for \textit{FedAvg} and \textit{FedAvgSch}, respectively. Without scheduling, smaller constellations did not surpass 60\% accuracy over three months. With scheduling and increased ground station coverage, convergence to 80\% accuracy was achieved, while reducing simulation times by up to 9X (from three months to roughly 10 days). These results highlight that increased access opportunities through scheduling, intra-cluster links, or larger ground station networks can transform previously impractical FL in space scenarios into feasible timeframes.

This analysis highlights several key lessons for designing FL algorithms for satellite constellations. Most algorithms achieve satisfactory accuracy if sufficient aggregation opportunities are available, with larger constellations and widespread ground station networks enabling performance above 80\% on FEMNIST. Achieving these guarantees, however, comes at the cost of longer training times. We next examine how round durations and satellite idle time impact overall training efficiency.

\subsection{Round Durations: The Constraint of Time}
\label{sec:roundurr}

Round duration is a critical bottleneck for FL in satellite constellations, directly affecting model convergence. While each constellation requires the same number of FL rounds for a given model and dataset, variations in communication windows and satellite availability can extend the time per round. Reducing these durations is essential to improve training efficiency and make space-based FL practical. Heatmaps of round durations across all experiments are shown in \Cref{fig:base_durations}, highlighting the influence of constellation design and communication infrastructure on training speed.

\begin{figure*}[htbp]
\centering
\begin{subfigure}{.3\textwidth}
  \centering
  \includegraphics[width=
0.96\linewidth]{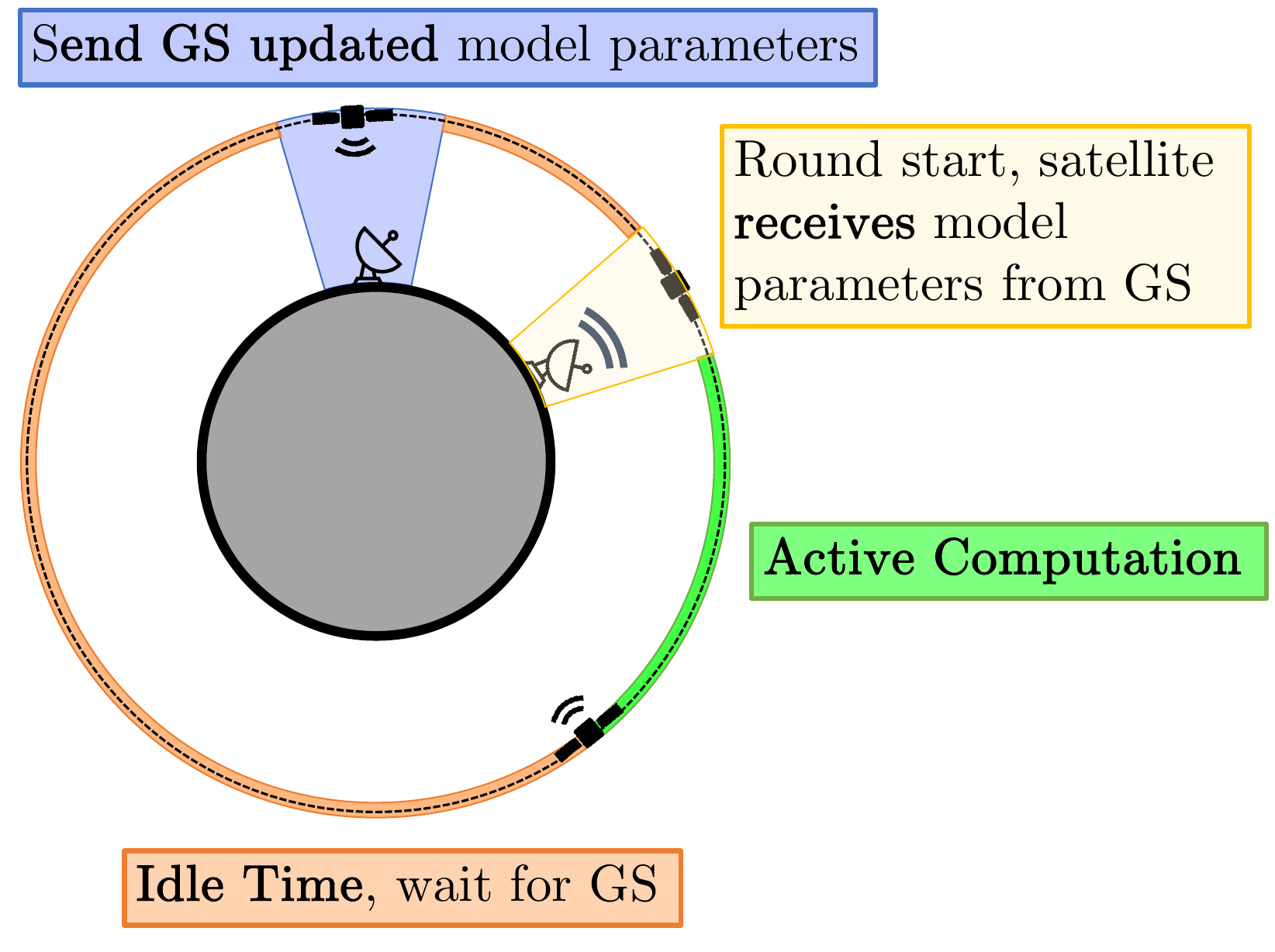}
  \caption{Idle times seen for a satellite in a \textit{FedAvg} FL aggregation setup.}
  \label{fig:sub1}
\end{subfigure}%
\hfill
\begin{subfigure}{.3\textwidth}
  \centering
  \includegraphics[width=\linewidth]{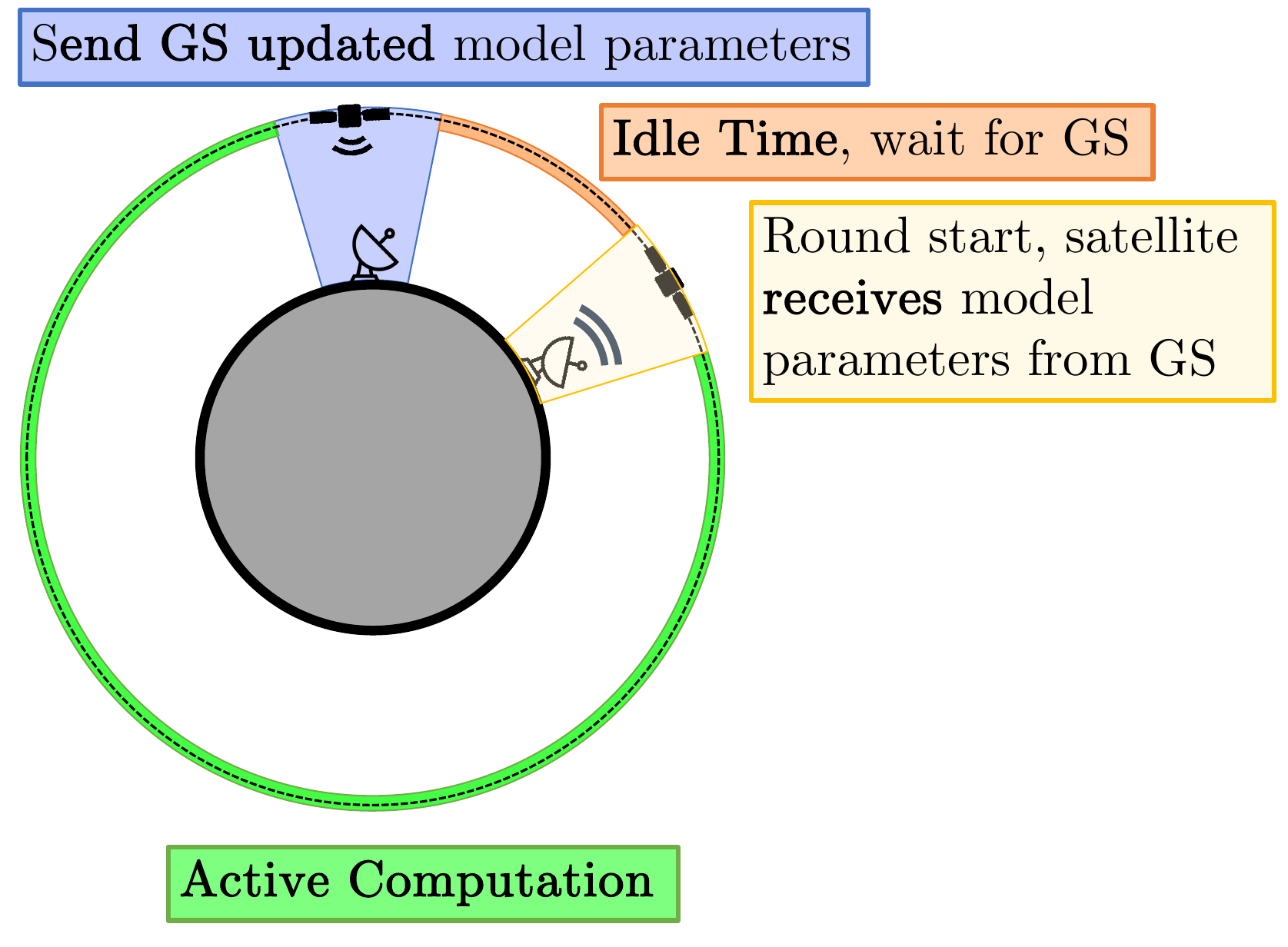}
  \caption{Idle times seen for a satellite in a \textit{FedProx} FL aggregation setup.}
  \label{fig:sub2}
\end{subfigure}
\hfill
\begin{subfigure}{.3\textwidth}
  \centering
  \includegraphics[width=\linewidth]{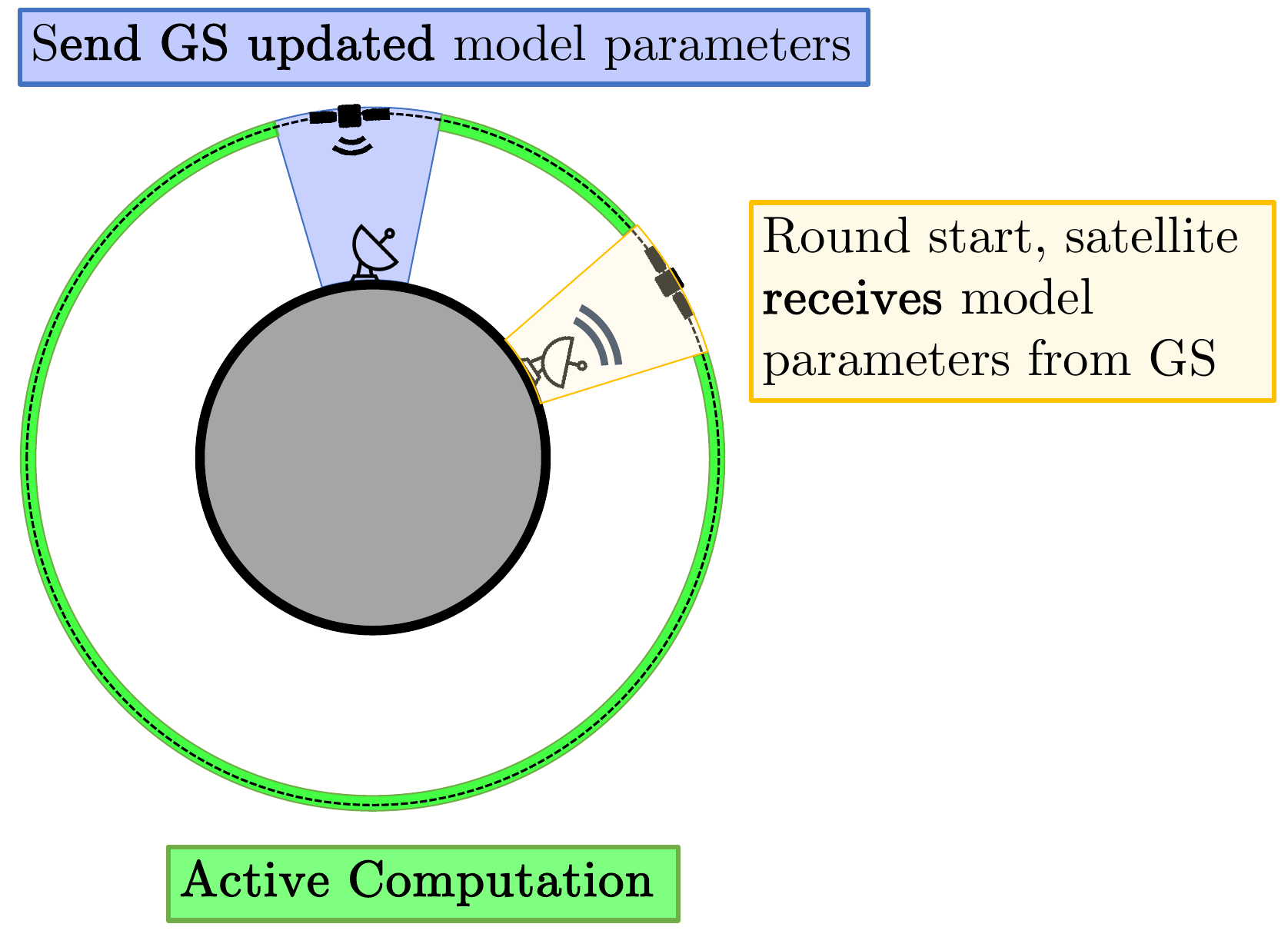}
  \caption{Idle times seen for a satellite in a \textit{FedBuff} FL aggregation setup.}
  \label{fig:sub3}
\end{subfigure}
\caption{Idle times for each aggregation method broken down along an example satellites orbit, with orange representing idle times, blue representing satellite to ground station communication times, yellow for ground station to satellite communication times, and green for active computation time on-board the satellite. \textit{FedBuff} has almost virtually no idle time in comparison to \textit{FedAvg}, which waits both in the model sending and receiving portion of the satellites orbit, and \textit{FedProx} which waits in the model receiving stage of the FL aggregation.}
\label{fig:idledescription}
\vspace{-.5em}
\end{figure*}

In this heatmap, each round aggregates at most 10 clients, including in asynchronous algorithms like \textit{FedBuff}. For smaller constellations, this means round durations can match those of constellations two or four times larger, ensuring fair comparisons. If fewer clients exist due to constellation size (e.g., 5 clusters with 1 satellite each), all satellites participate in every aggregation round.

Round durations correlate strongly with the number of ground stations. Increasing the network from 1 to 3 or 3 to 5 stations significantly reduces durations by enabling earlier aggregation. Beyond five ground stations, however, reductions plateau, with minimal improvement observed for 10 or 13 stations. This reflects diminishing returns from redundant ground stations and the geographic deployment of facilities (e.g., Tromsø’s polar location primarily benefits polar-orbiting satellites). Optimal satellite-to-ground station placements, rather than sheer quantity, could be explored to maximize access opportunities. By leveraging constant intra-cluster communication, a small number of ground stations may effectively cover nearly all communications.

Scheduling consistently reduces round durations across \textit{FedAvgSch}, \textit{FedProxSch}, and \textit{FedProxSchV2}, particularly for larger clusters. Increasing satellites per cluster, rather than the number of clusters, also shortens durations. This is due to cluster-specific “trailing effects”: satellites in the same orbit can access a ground station sequentially with minimal delay, whereas satellites spread across orbits must contact the same station at different times, lengthening rounds. For constellation designers, these results indicate that adding satellites to existing clusters is more effective than increasing cluster count, especially when leveraging ISL links. Fewer clusters with more satellites improve round durations, while scheduling only reduces durations when sufficient access windows exist, which is a condition more reliably met with intra-cluster communications. 

In these explorations, we find that the \textbf{speedups in FL round durations are most significant using larger satellite constellations} as they have numerous access windows available to optimize for in client selection. There are \textit{limits to how much one can optimize for time spent for model aggregation over a small network}, which will be an important satellite constellation design consideration for future FL researchers.

\begin{figure*}[!ht]

    \begin{subfigure}{\textwidth}
    \centering
    \includegraphics[trim={0 1cm 0 0},clip,width=\textwidth]{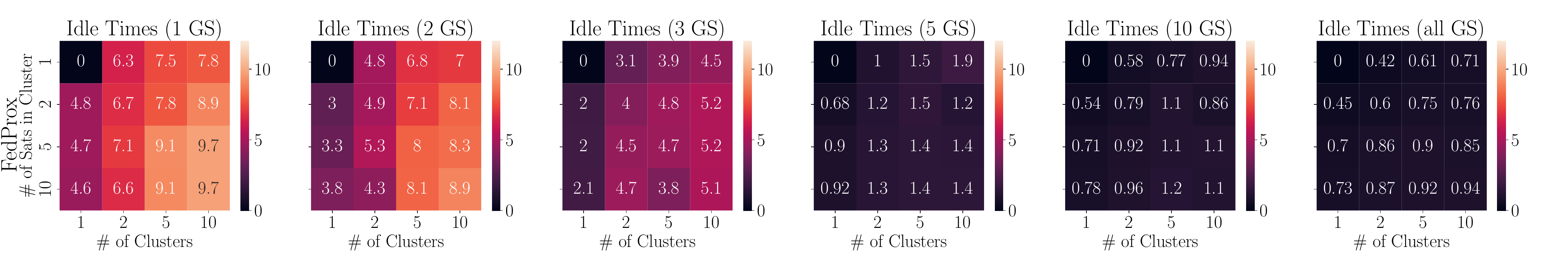}
    \label{fedProx:iavg}
    \end{subfigure}

    \begin{subfigure}{\textwidth}
    \centering
    \includegraphics[trim={0 1cm 0 0},clip,width=\textwidth]{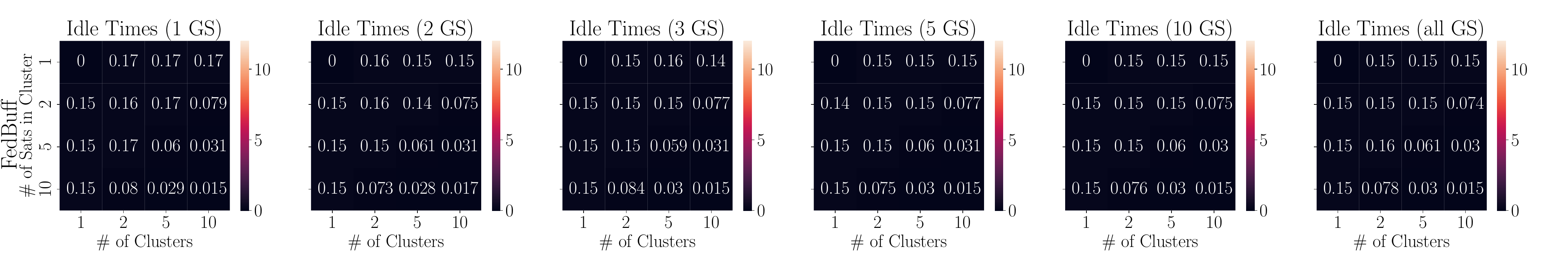}
    \caption{Baseline FL Algorithms, with no Scheduling or Intra Satellite Communications}
    \vspace{0.5em}
    \label{fedBuff:iavg}
    \end{subfigure}

    \begin{subfigure}{\textwidth}
    \centering
    \includegraphics[trim={0 1cm 0 0},clip,width=\textwidth]{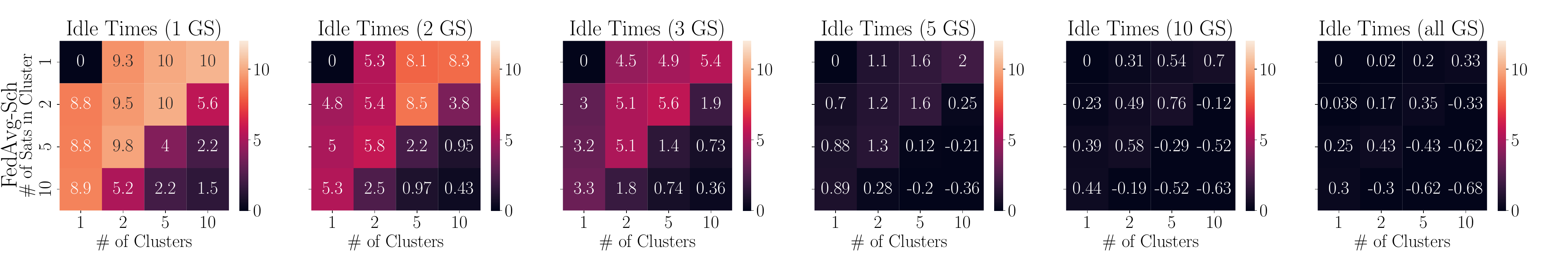}
    \label{fedavg2_10:iavg}
    \end{subfigure}

    \begin{subfigure}{\textwidth}
    \centering
    \includegraphics[trim={0 1cm 0 0},clip,width=\textwidth]{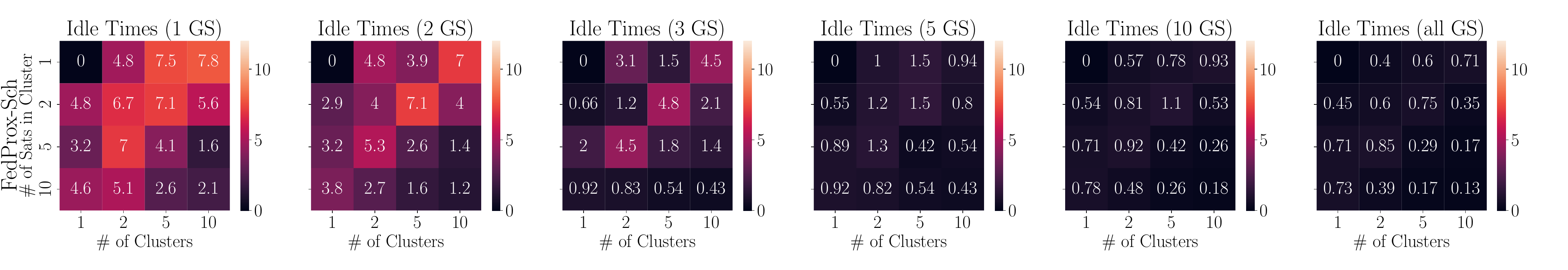}
    \label{fedProx2:iavg}
    \end{subfigure}

    \begin{subfigure}{\textwidth}
    \centering
    \includegraphics[trim={0 1cm 0 0},clip,width=\textwidth]{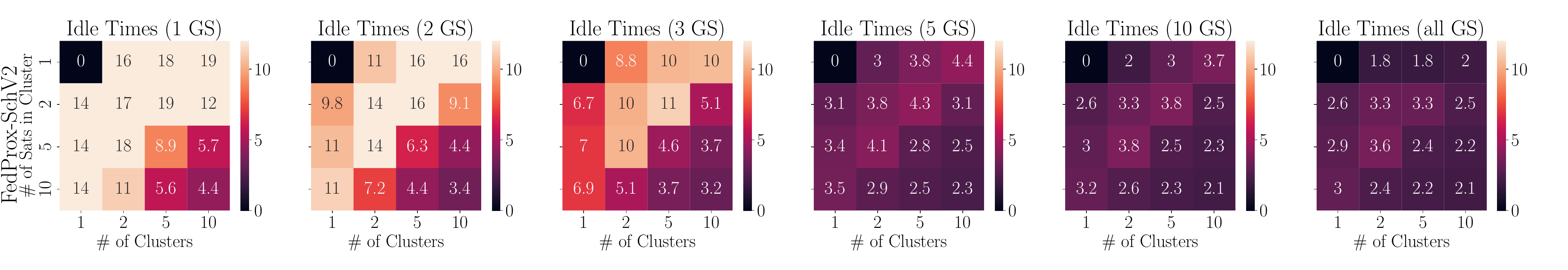}
    \caption{FL Algorithms with a Scheduler, but without Intra Satellite Communications}
    \vspace{0.5em}
    \label{fedProx22:iavg}
    \end{subfigure}
    
    \begin{subfigure}{\textwidth}
    \centering
    \includegraphics[trim={0 0cm 0 0cm},clip,width=\textwidth]{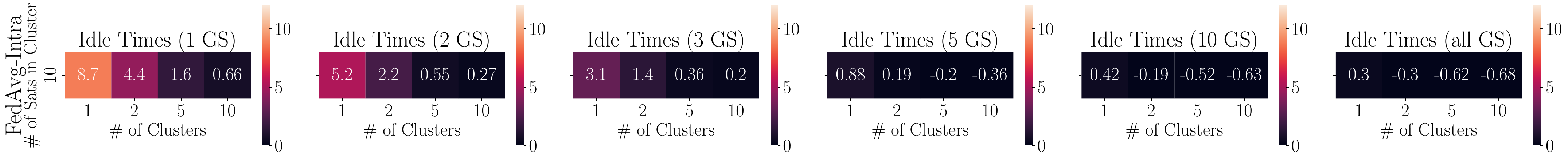}
    \label{fedavg3:idle}
    \end{subfigure}

    \begin{subfigure}{\textwidth}
    \centering
    \includegraphics[trim={0 0cm 0 0cm},clip,width=\textwidth]{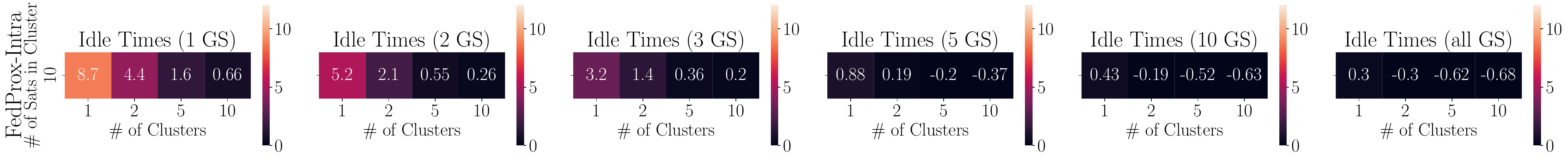}
    \caption{FL Algorithms with both a Scheduler, and enabled Intra Satellite Communications}
    \label{fedprox3:idle}
    \end{subfigure}
    
    \caption{Heatmaps depicting the satellite idle time per hour during round aggregation, testing changes in the \#s of clusters, \#s of satellites per cluster, and \#s of ground stations available to connect in the network. \textit{FedAvg}, \textit{FedProx}, and \textit{FedBuff}, are tested measured against versions of the same algorithm but with scheduled and Intra SL enabled communications. Again, the number of ground stations seems to have the largest impact on round duration length, but a point of interest is the drop in idle time through \textit{FedBuff}, which performs computation continously throughout each satellte's orbit.}\label{fig:idleavg_normal}
\end{figure*}

\subsection{Satellite Idle Time: Eliminating the Gaps}
\label{subsec:idletime}
To fully understand how each FL round is spent on individual satellites, we evaluate idle time, which we define as periods when satellites are neither computing nor communicating. Idle time varies considerably across algorithms due to differences in communication and aggregation strategies, as illustrated in \Cref{fig:idledescription}.

For the baseline algorithms, \textit{FedAvg}, \textit{FedProx}, and \textit{FedBuff}, idle time arises from distinct processes. In \textit{FedAvg}, satellites wait to receive the global model, perform training for a fixed number of epochs, and then wait again to transfer updates to a ground station; model transfers can take several minutes depending on model size. \textit{FedProx} similarly incurs initial waiting time, but satellites continue local training until reaching a ground station for communication, resulting in less idle time compared to \textit{FedAvg}. \textit{FedBuff}, the only asynchronous method, experiences minimal idle time, limited primarily to the model transfer process.

\Cref{fig:idleavg_normal} outlines the extensive heatmaps sweeping the three constellation parameters $\#$ of clusters, $\#$ of satellites per cluster, and $\#$ of ground stations. The metrics are specific to per-satellite idle times for each FL algorithm to enable fair comparisons across constellations with varying client counts. For \textit{FedAvg}, idle time is primarily driven by ground station availability, with reductions plateauing beyond five stations. In \textit{FedProx}, smaller constellations ($\leq10$ clients per round) show lower idle times, as all satellites participate continuously. \textit{FedBuff}, due to its asynchronous design, consistently minimizes idle time, making it the most effective method despite longer round durations. Applying scheduling further reduces idle time, particularly in large constellations where many clients may not participate in every round. The magnitude of improvement depends on the underlying FL algorithm, as structural differences in aggregation and computation protocols set the baseline for idle time.

An important consideration is that the round duration for each scenario inherently limits the maximum idle time on each satellite. Regardless of how much idle time is optimized, a larger reduction in total FL training time can be achieved by shortening the overall duration of each aggregation round, as idle time scales with round length. \textbf{Prioritizing round duration reductions}, rather than maximizing continuous computation on each satellite, enables FL to be executed on more practical mission timescales.








\section{Conclusions}
This work brings federated learning into space in earnest, moving beyond prior conceptual or feasibility studies by implementing, evaluating, and providing tools for operational space FL. We introduce a modular \textit{space-ification framework} that adapts any terrestrial FL algorithm to orbital conditions, including client selection, round completion, and model evaluation. We demonstrate this framework with \textit{FedAvg}, \textit{FedProx}, and \textit{FedBuff}. To improve performance in orbital environments, we develop constellation-specific enhancements, specifically scheduling optimizations and intra-satellite communications, that are applicable to any space-ified FL algorithm, not just tailored to a single method. These enhancements reduce round durations, minimize idle time, and improve resource utilization, making FL more practical for large satellite constellations. Finally, we evaluate these methods through extensive simulations across 768 configurations of satellite clusters, satellite counts, and ground station networks. Using this framework, all adapted algorithms achieve over 80\% accuracy on standard benchmarks, and with the proposed enhancements, training times for a 100-satellite constellation can be reduced by up to 9X (from three months to roughly 10 days).

From these experiments, several key design lessons emerge for future satellite constellations leveraging on-orbit FL. First, balanced scheduling is essential to prevent overly aggressive reductions in per-round computation, ensuring sufficient local training per satellite. Second, increasing access points, via additional ground stations or intra-cluster links, reduces FL round durations and improves resource utilization. Third, ground station placement exhibits diminishing returns beyond a certain network size; prioritizing coverage in key geographic locations is more effective than sheer quantity. Finally, cluster design matters: increasing satellites per cluster leverages intra-satellite links to reduce round durations more effectively than simply adding clusters.

These insights provide a practical roadmap for researchers and satellite operators, showing that careful constellation design, scheduling, and communication planning can enable scalable, collaborative FL in orbit. By bridging algorithmic innovation with constellation and communication design, this work establishes a foundation for operational on-orbit learning across Earth-orbiting networks.

\acknowledgements 
This research was supported by the Hertz Foundation and the National Defense Science and Engineering Graduate (NDSEG) Fellowship Program.


\bibliography{references}
\bibliographystyle{IEEEtran}

\thebiography

\begin{biographywithpic}{Grace Ra Kim}{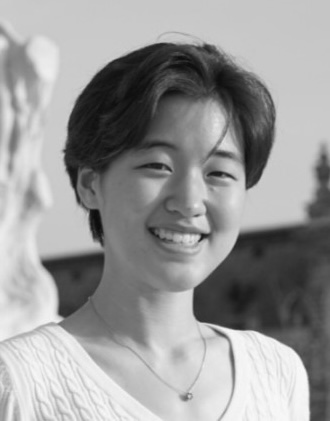}
is a PhD student in the Stanford Intelligent Systems Laboratory (SISL) in the department of Aeronautics and Astronautics at Stanford University, as a Hertz and NDSEG Fellow. She received her M.Phil. in Advanced Computer Science at the University of Cambridge as a Marshall Scholar in 2024, and graduated cum laude with high honors in a B.S. in Engineering Sciences from Harvard University in 2023. Her research interests lie at the intersection of space traffic management, safe autonomous systems in space, and space policy.
\end{biographywithpic} 

\begin{biographywithpic}{Filip Svoboda}{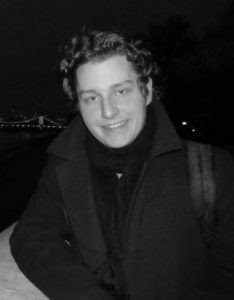} is a researcher working on deep learning efficiency in embedded systems, affiliated with the Oxford Machine Learning Systems Lab and the Autonomous, Intelligent Machines and Systems centre at the University of Oxford. His research focuses on resource-constrained inference for embedded and wearable devices and developing the Rational Automated Machine Learner paradigm, which incorporates resource costs and preferences into automated machine learning. He is also co-founder and director of the Cambridge Neural Group, which promotes deep learning accessibility in the UK and abroad.
\end{biographywithpic}

\begin{biographywithpic}{Nicholas D. Lane}{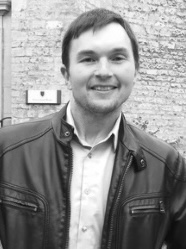} is a Professor in the Department of Computer Science and Technology at the University of Cambridge, where he holds a Royal Academy of Engineering Chair in De-centralized AI. He leads the Cambridge Machine Learning Systems lab (CaMLSys) focused on pioneering scalable, robust machine learning systems. Lane is co-founder and Chief Scientific Officer of Flower Labs, a venture-backed AI company. He has received multiple best paper awards, including ACM/IEEE IPSN and ACM UbiComp, as well as the ACM SIGMOBILE Rockstar Award for his work enabling mobile devices to understand and react to complex user behaviors through novel learning algorithms and system design.

\end{biographywithpic}

\end{document}